%% file: neurips_2025.tex
\documentclass{article}

\PassOptionsToPackage{numbers, compress}{natbib}

\usepackage[final]{neurips_2025}

\newcommand{\best}[1]{\textbf{#1}} 

\usepackage{pifont}
\usepackage[utf8]{inputenc} 
\usepackage[T1]{fontenc}    
\usepackage{hyperref}       
\usepackage{url}            
\usepackage{booktabs}       
\usepackage{amsfonts}       
\usepackage{nicefrac}       
\usepackage{pdfpages}
\usepackage{microtype}      
\usepackage{xcolor}         
\usepackage{lineno}
\usepackage{graphicx}
\usepackage{multirow, makecell}
\usepackage{caption}
\usepackage[table]{xcolor}
\usepackage{wrapfig}
\usepackage{amsmath} 
\usepackage{enumitem}

\newcommand{\second}[1]{\underline{{#1}}}

\newcommand{\pcite}[1]{\textcolor{red}{[cite related works]}}

\title{Toward a Vision-Language Foundation Model for Medical Data: Multimodal Dataset and Benchmarks for Vietnamese PET/CT Report Generation}

\author{
\textbf{Huu Tien Nguyen}$^{1}$\thanks{Equal contribution.} \quad
\textbf{Dac Thai Nguyen}$^{1*}$ \quad
\textbf{The Minh Duc Nguyen}$^{1}$ \quad \\ 
\textbf{Trung Thanh Nguyen}$^{2\dagger}$ \quad
\textbf{Thao Nguyen Truong}$^{3}$ \quad
\textbf{Huy Hieu Pham}$^{4}$ \quad 
\textbf{Johan Barthelemy}$^{5}$ \\
\textbf{Minh Quan Tran}$^{5}$ \quad
\textbf{Thanh Tam Nguyen}$^{6}$ \quad
\textbf{Quoc Viet Hung Nguyen}$^{6}$ \\ 
\textbf{Quynh Anh Chau}$^{7}$ \quad 
\textbf{Hong Son Mai}$^{8}$ \quad 
\textbf{Thanh Trung Nguyen}$^{8}$ \quad
\textbf{Phi Le Nguyen}$^{1}$\thanks{Corresponding authors.}   \\
$^1$AI4LIFE, Hanoi University of Science and Technology, Vietnam \quad
$^2$Nagoya University, Japan \\ 
$^3$AIST, Japan \quad
$^4$VinUniversity, Vietnam \quad 
$^5$NVIDIA, USA \quad
$^6$Griffith University, Australia \\
$^7$Hanoi Medical University, Vietnam \quad
$^8$108 Military Central Hospital, Vietnam 
\\
\footnotesize\texttt{\{tien.nh205033, thai.nd242110m, duc.ntm204904\}@sis.hust.edu.vn} \\
\footnotesize\texttt{nguyent@cs.is.i.nagoya-u.ac.jp,
nguyen.truong@aist.go.jp} \\ 
\footnotesize\texttt{hieu.ph@vinuni.edu.vn,
\{jbarthelemy, tranminhq\}@nvidia.com} \\
\footnotesize\texttt{\{henry.nguyen, t.nguyen19\}@griffith.edu.au, alex.hong.son@gmail.com} \\
\footnotesize\texttt{05230802@daihocyhanoi.edu.vn,
trungntc10@benhvien108.vn, lenp@soict.hust.edu.vn}
}

\begin{document}

\maketitle
\begin{abstract}
Vision-Language Foundation Models (VLMs), trained on large-scale multimodal datasets, have driven significant advances in Artificial Intelligence (AI) by enabling rich cross-modal reasoning. Despite their success in general domains, applying these models to medical imaging remains challenging due to the limited availability of diverse imaging modalities and multilingual clinical data. Most existing medical VLMs are trained on a subset of imaging modalities and focus primarily on high-resource languages, thus limiting their generalizability and clinical utility. To address these limitations, we introduce a novel Vietnamese-language multimodal medical dataset consisting of $2{,}757$ whole-body PET/CT volumes from independent patients and their corresponding full-length clinical reports. This dataset is designed to fill two pressing gaps in medical AI development: (1) the lack of PET/CT imaging data in existing VLMs training corpora, which hinders the development of models capable of handling functional imaging tasks; and (2) the underrepresentation of low-resource languages, particularly the Vietnamese language, in medical vision-language research. To the best of our knowledge, this is the first dataset to provide comprehensive PET/CT-report pairs in Vietnamese. We further introduce a training framework to enhance VLMs' learning, including data augmentation and expert-validated test sets. We conduct comprehensive experiments benchmarking state-of-the-art VLMs on downstream tasks, including medical report generation and visual question answering. The experimental results show that incorporating our dataset significantly improves the performance of existing VLMs. However, despite these advancements, the models still underperform on clinically critical criteria, particularly the diagnosis of lung cancer, indicating substantial room for future improvement. We believe this dataset and benchmark will serve as a pivotal step in advancing the development of more robust VLMs for medical imaging, particularly in low-resource languages, and improving their clinical relevance in Vietnamese healthcare.
\end{abstract}

\section{Introduction}
\input{tex/1-introduction}


\section{ViMed-PET: The proposed Vietnamese Vision-Language Medical Dataset of PET/CT Images and Clinical Reports}
\label{our_dataset}
\input{tex/3-1-dataset}

\section{Model Selection and Fine-tuning Flow}
\input{tex/3-benchmarks}
\section{Evaluation Results}
\input{tex/4-evaluation_results}

\section{Conclusion}
\input{tex/5-conclusion}



\bibliographystyle{unsrtnat} 
\bibliography{references}

\newpage
\clearpage
\section*{NeurIPS Paper Checklist}

\begin{enumerate}
\item {\bf Claims}
    \item[] Question: Do the main claims made in the abstract and introduction accurately reflect the paper's contributions and scope?
    \item[] Answer: \answerYes{} 
    \item[] Justification: We have already outlined this in the Abstract and Introduction.

\item {\bf Limitations}
    \item[] Question: Does the paper discuss the limitations of the work performed by the authors?
    \item[] Answer: \answerYes{} 
    \item[] Justification: The limitations of this study are presented in Section 5.

\item {\bf Theory assumptions and proofs}
    \item[] Question: For each theoretical result, does the paper provide the full set of assumptions and a complete (and correct) proof?
    \item[] Answer: \answerNA{} 
    \item[] Justification: The paper does not include theoretical results.

    \item {\bf Experimental result reproducibility}
    \item[] Question: Does the paper fully disclose all the information needed to reproduce the main experimental results of the paper to the extent that it affects the main claims and/or conclusions of the paper (regardless of whether the code and data are provided or not)?
    \item[] Answer: \answerYes{} 
    \item[] Justification: 
    We provided detailed descriptions of the training data, network architecture,
    and experimental details to ensure the reproducibility of the results.

\item {\bf Open access to data and code}
    \item[] Question: Does the paper provide open access to the data and code, with sufficient instructions to faithfully reproduce the main experimental results, as described in supplemental material?
    \item[] Answer: \answerYes{} 
    \item[] Justification: We provide the public URL to our dataset, and the source code is submitted along with this paper.

\item {\bf Experimental setting/details}
    \item[] Question: Does the paper specify all the training and test details (e.g., data splits, hyperparameters, how they were chosen, type of optimizer, etc.) necessary to understand the results?
    \item[] Answer: \answerYes{} 
    \item[] Justification: We describe the data splitting in the main text, while the hyperparameters used for the experiments are detailed in the Appendix.

\item {\bf Experiment statistical significance}
    \item[] Question: Does the paper report error bars suitably and correctly defined or other appropriate information about the statistical significance of the experiments?
    \item[] Answer: \answerNo{} 
    \item[] Justification: We do not report the error bar due to the high cost for training LLMs.

\item {\bf Experiments compute resources}
    \item[] Question: For each experiment, does the paper provide sufficient information on the computer resources (type of compute workers, memory, time of execution) needed to reproduce the experiments?
    \item[] Answer: \answerYes{} 
    \item[] Justification: Each experiment was conducted on a machine with four A100 GPUs, with training and evaluation completed within three days (see Appendix for details).
    
\item {\bf Code of ethics}
    \item[] Question: Does the research conducted in the paper conform, in every respect, with the NeurIPS Code of Ethics \url{https://neurips.cc/public/EthicsGuidelines}?
    \item[] Answer: \answerYes{} 
    \item[] Justification: This study complies with the NeurIPS Code of Ethics.

\item {\bf Broader impacts}
    \item[] Question: Does the paper discuss both potential positive societal impacts and negative societal impacts of the work performed?
    \item[] Answer: \answerNo{} 
    \item[] Justification: This study aims to benchmark existing LLMs on a medical dataset and does not present any negative societal impact.
    
\item {\bf Safeguards}
    \item[] Question: Does the paper describe safeguards that have been put in place for responsible release of data or models that have a high risk for misuse (e.g., pretrained language models, image generators, or scraped datasets)?
    \item[] Answer: \answerNA{} 
    \item[] Justification: 

\item {\bf Licenses for existing assets}
    \item[] Question: Are the creators or original owners of assets (e.g., code, data, models), used in the paper, properly credited and are the license and terms of use explicitly mentioned and properly respected?
    \item[] Answer: \answerYes{} 
    \item[] Justification: We indicate licenses of public datasets in Appendix.

\item {\bf New assets}
    \item[] Question: Are new assets introduced in the paper well documented and is the documentation provided alongside the assets?
    \item[] Answer: \answerYes{}{} 
    \item[] Justification: We provide the documentation along with the code.

\item {\bf Crowdsourcing and research with human subjects}
    \item[] Question: For crowdsourcing experiments and research with human subjects, does the paper include the full text of instructions given to participants and screenshots, if applicable, as well as details about compensation (if any)? 
    \item[] Answer: \answerNA{} 
    \item[] Justification: \answerNA{}

\item {\bf Institutional review board (IRB) approvals or equivalent for research with human subjects}
    \item[] Question: Does the paper describe potential risks incurred by study participants, whether such risks were disclosed to the subjects, and whether Institutional Review Board (IRB) approvals (or an equivalent approval/review based on the requirements of your country or institution) were obtained?
    \item[] Answer: \answerNo{} 
    \item[] Justification: We remove the private information from the dataset.

\item {\bf Declaration of LLM usage}
    \item[] Question: Does the paper describe the usage of LLMs if it is an important, original, or non-standard component of the core methods in this research? Note that if the LLM is used only for writing, editing, or formatting purposes and does not impact the core methodology, scientific rigorousness, or originality of the research, declaration is not required.
    \item[] Answer: \answerYes{}{} 
    \item[] Justification:  Our paper uses GPT-4o to generate augmented datasets based on a few-shot examples created by domain experts. We also leverage GPT-4o to post-process model outputs for evaluating clinical metrics. In addition, we utilize pretrained Mistral and LLaMA-2 models, combined with vision encoders, to conduct experiments on our dataset.
\end{enumerate}


\newpage
\appendix
\section{Medical Vision-Language Models}
\input{tex/2-related_work}

\section{Technical Appendices}
\input{tex/6-appendix}

\end{document}

%% file: tex/1-introduction.tex
Vision-Language Foundation Models (VLMs) have rapidly evolved as a cornerstone of modern Artificial Intelligence (AI), capable of jointly modeling information across visual and textual modalities. 
These models are typically pre-trained on diverse datasets encompassing billions of image-text pairs, enabling them to acquire generalized and transferable representations~\cite{radford2021learning, yang2023dawn, li2024llava, bai2025qwen2}. 
This cross-modal alignment allows VLMs to bridge the semantic gap between images and language, facilitating downstream tasks such as image captioning~\cite{bai2023qwen, liu2023visual, chen2025q}, visual question answering~\cite{li2024llava, bai2025qwen2, bai2023qwen, liu2023visual}, report generation~\cite{yan2024ahive, pham2024fg}, and even zero-shot image classification~\cite{radford2021learning, javed2024cplip}, with minimal task-specific supervision.

While VLMs such as CLIP~\cite{radford2021learning}, Flamingo~\cite{alayrac2022flamingo}, and GPT-4o~\cite{hurst2024gpt}  have demonstrated exceptional performance on natural image benchmarks, transferring this success to the medical domain remains an ongoing challenge. 
The primary barrier lies in the domain shift: medical images fundamentally differ from natural images in terms of texture, structure, semantics, and purpose~\cite{nori2023can, moon2022multi}. 
Furthermore, the textual annotations associated with medical images are typically richer, more technical, and context-sensitive, often demanding expert-level knowledge for accurate interpretation~\cite{Annotations, davis2025knowledge, KnowledgeMatters}.

In response to these limitations, recent efforts have focused on developing medical-specific VLMs, including MedCLIP~\cite{MedCLIP2021} and MedFlamingo~\cite{Med-Flamingo}.
These models aim to adapt general-purpose VLMs' architecture to the medical domain by retraining or fine-tuning on domain-specific datasets. 
However, existing models are still constrained in several critical aspects. 
\textit{First,} the visual modality coverage in current medical VLMs is narrow. 
Most existing works focus on well-established imaging types such as chest X-rays~\cite{chambon2022roentgen}, Magnetic Resonance Imaging (MRI) and Computed Tomography (CT) scans~\cite{xu2024medvilam}, and histopathology slides~\cite{xin2025med3dvlm}. 
In contrast, functional imaging modalities such as Positron Emission Tomography (PET), which are essential in oncology, cardiology, and neurology, remain significantly underrepresented in current datasets and VLMs.
\textit{Second,} the linguistic side of existing VLMs, and vision-language datasets, is largely monolingual, overwhelmingly dominated by English~\cite{nguyen2024multilingual}. 
Very few resources consider linguistic inclusivity, resulting in low-resource languages such as Vietnamese being severely underrepresented in developing and evaluating VLMs.
In addition, most existing datasets include only brief image captions~\cite{lin2023pmc, ruckert2024rocov2} or limited diagnostic annotations~\cite{johnson2019mimic}, which are insufficient to fully capture the complex and nuanced information embedded in medical images.

\begin{wraptable}{r}{0.50\textwidth}
\vspace{-13pt}
\caption{Limitations of current VLMs in Vietnamese reports generating from PET images. R-1 and R-L denote ROUGE-1 and ROUGE-L scores. 
\textit{*GPT-4o is evaluated under few-shot prompting.}
}
\label{tab:preliminary_pet_results}
\centering
\setlength\tabcolsep{3pt} 
\resizebox{1.0\linewidth}{!}{
\begin{tabular}{l|c|c|c|c}
\toprule
\textbf{Model} & \textbf{BLEU-4} $\uparrow$ & \textbf{R-1} $\uparrow$ & \textbf{R-L} $\uparrow$ & \textbf{BERT} $\uparrow$ \\
\midrule
LLaVA-Med~\cite{LLaVAMed2023} &\hspace{2pt} 0.01 & 50.08 & 27.89 & 64.63 \\
M3D~\cite{M3DLaMed2022}           & \hspace{2pt} 0.04 & 41.01 & 23.53 & 67.21 \\
RadFM~\cite{RadFM2022}            & \hspace{2pt} 0.06 & 54.23 & 28.33 & 69.49 \\
GPT-4o*~\cite{hurst2024gpt}    & 31.12 & 67.96 & 52.76 & 81.09 \\
\bottomrule
\end{tabular}
}
\end{wraptable}
Our preliminary experiments highlight significant limitations of current medical VLMs when applied to PET/CT imaging data. Table~\ref{tab:preliminary_pet_results} presents the results of a Vietnamese clinical report generation task using PET/CT images from a dataset we curated, comprising 1,725 clinical case studies. As shown, models such as LLaVA-Med~\cite{LLaVAMed2023}, M3D~\cite{M3DLaMed2022}, and RadFM~\cite{RadFM2022} yield near-zero BLEU-4 scores and low ROUGE and BERT score metrics, reflecting their poor capacity to generate coherent and clinically relevant reports. While GPT-4o~\cite{hurst2024gpt} demonstrates relatively better performance, its BLEU-4 score remains at a modest 31\%, indicating inadequate generation quality.
These findings underscore the pressing need for more diverse training data regarding imaging modalities and linguistic representation to enhance the robustness and generalizability of medical VLMs.

To address this challenge, we introduce the first large-scale paired dataset of PET/CT images and corresponding clinical reports in Vietnamese, a language with limited medical AI resources. 
Our focus is motivated by these factors:
\begin{itemize}[leftmargin=*, itemsep=1pt ,topsep=1pt]
\item \textbf{Clinical significance:} PET/CT scan is indispensable in modern diagnostic workflows, especially in oncology, where it enables non-invasive assessments of tumor metabolism and spread \cite{kim2016non, lewis2015imaging, gambhir2002molecular, schwenck2023advances}. 
Its importance in early diagnosis, staging, and treatment monitoring is unparalleled, yet it remains underutilized in AI due to data scarcity.
\item \textbf{Data scarcity and accessibility:} Public PET/CT datasets are rare. To our knowledge, no existing dataset offers paired PET/CT images with detailed clinical reports.  
PET/CT scans are also among the most expensive imaging procedures~\cite{alberts2025long, naghavi2023cost}, further limiting data availability and open access.
\item \textbf{Language equity:} The lack of medical image-report datasets in Vietnamese exacerbates health data inequity. With over 100 million native speakers\footnote{Explore the World Population Through Data (2025).
\url{https://worldpopulationreview.com/countries/vietnam}. Accessed on 2025/12/01.}, Vietnamese represents a substantial user base that remains excluded from AI-enabled healthcare technologies.
\end{itemize}

The main contributions of this study are as follows:
\begin{enumerate}[leftmargin=*,topsep=1pt, itemsep=1pt]
    \item We introduce a comprehensive dataset comprising 2,757 whole-body PET/CT volumes from independent patients along with full-length Vietnamese clinical reports.
    The dataset spans a demographically and pathologically diverse patient population, reflecting real-world clinical variability.
    It provides a valuable resource for advancing the training of medical VLMs, with the potential to support a broader range of 
    modalities and enable multilingual development, especially for low-resource languages.

    \item We develop a data augmentation framework that enriches the visual and textual components of the dataset, improving its effectiveness for model training and generalization.
    
    \item We leverage our newly curated PET/CT image-report dataset, named ViMed-PET, to fine-tune state-of-the-art VLMs and evaluate their performance on tasks such as medical report generation and visual question answering.
    Experimental results show notable gains, enhancing the capabilities of 
    pre-trained medical VLMs.

    \item We collaborate with medical domain experts to develop a clinically validated test set specifically tailored for lung cancer diagnosis.
    This test set incorporates structured, clinically relevant evaluation metrics that assess model performance in real-world diagnostic scenarios.
    Rather than relying solely on conventional Natural Language Processing metrics that emphasize lexical matching, our benchmark provides a comprehensive evaluation of the ability of a model to address the nuanced and complex demands of clinical lung cancer diagnosis. 
    This offers a more holistic and meaningful assessment of medical VLMs' effectiveness.
\end{enumerate}


%% file: tex/3-1-dataset.tex
\begin{table}[t]
\caption{Comparison of our ViMed-PET dataset with existing medical vision-language datasets.
Our dataset is the first large-scale PET/CT dataset with clinical reports in Vietnamese, stored in standard medical DICOM format.
 \textit{``Multiple''} indicates more than one modality (e.g., CT, MRI, X-ray). 
 \textit{``K''} is thousand. 
 \textit{(*)} Values show the number of 2D slices/images extracted from 3D volumes.}
 \vspace{2pt}
\label{tab:dataset_comparison}
\centering
\resizebox{\textwidth}{!}{
\begin{tabular}{l|c|c|c|c|c|r|r}
\toprule
\multirowcell{2}[-2pt][l]{\textbf{Dataset Name}} & \multicolumn{2}{c|}{\textbf{Text}} & \multicolumn{3}{c|}{\textbf{Image}} & \multicolumn{2}{c}{\textbf{Modality Size}} \\
\cmidrule(lr){2-3} \cmidrule(lr){4-6} \cmidrule(lr){7-8}
 & PET-related & {Type} & {3D Volume} & {PET/CT} &  {Others} & \multicolumn{1}{c|}{{PET/CT}} & \multicolumn{1}{c}{{Others}} \\
\midrule
MIMIC-CXR~\cite{johnson2019mimic}        & \ding{55}       &  Report   &  \ding{55}  & \ding{55} & X-ray                   &   \multicolumn{1}{c|}{--}   & 227K  \\
PMC-OA~\cite{lin2023pmc}                 & \ding{55}      &  Caption           &  \ding{55}   & \ding{51}    &  Multiple          & 600K & 1,646K \\
ROCOv2~\cite{ruckert2024rocov2}          &    \ding{55}     &  Caption           &  \ding{55} & \ding{51}  &  Multiple          & 432 & 79K \\
\midrule
CT-RATE~\cite{Ct2rep}                    & \ding{55}       & Report              &  \ding{51}  & \ding{55}  & CT           & \multicolumn{1}{c|}{--} & 50K  \\
M3D-Data~\cite{M3DLaMed2022}             &    \ding{55}       &  Report              &  \ding{51}  & \ding{55}  & CT           & \multicolumn{1}{c|}{--} & 120K \\
MedMD-3D~\cite{RadFM2022}                &  \ding{55}       & Caption/Report            & \ding{51}  & \ding{55}  & Multiple  & \multicolumn{1}{c|}{--} & 500K \\
\midrule
RIDER Lung PET-CT~\cite{muzi2015riderlungpetct}         & \ding{55} & --        & \ding{51} & \ding{51} & -- & 266K (*) & \multicolumn{1}{c}{--} \\
Head-Neck PET-CT~\cite{vallieres2017headneckpetct}      & \ding{55} & --        & \ding{51} & \ding{51} & -- & 123K (*) & \multicolumn{1}{c}{--} \\
Lung-PET-CT-Dx~\cite{li2020lungpetctdx}                 & \ding{55} & --        & \ding{51} & \ding{51} & -- & 251K (*) & \multicolumn{1}{c}{--} \\
FDG-PET-CT-Lesions~\cite{gatidis2022fdgpetctlesions}    & \ding{55} & --        & \ding{51} & \ding{51} & -- & 917K (*) & \multicolumn{1}{c}{--} \\
\midrule
\midrule
\textbf{Our ViMed-PET Dataset} & \ding{51} & Report & \ding{51} & \ding{51} & \multicolumn{1}{c|}{--} & 1,567K  (*) & \multicolumn{1}{c}{--} \\
\bottomrule
\end{tabular}
}
\begin{minipage}{\linewidth}
\vspace{2.5pt}
\centering
\footnotesize
\vspace{3pt}
\footnotesize{\textit{For our ViMed-PET dataset, ${1,567}$K paired slices correspond to 2,757 paired whole-body PET/CT volumes.}}
\end{minipage}

\end{table}

\subsection{Existing Medical Multimodal Datasets}
Recent advances in medical VLMs have been driven by datasets that align medical images with associated textual annotations.
Table~\ref{tab:dataset_comparison} summarizes existing datasets, which primarily cover CT, MRI, and X-ray modalities and provide either image captions or diagnostic reports. For 2D imaging, representative examples include MIMIC-CXR~\cite{johnson2019mimic}, PMC-OA~\cite{lin2023pmc}, and ROCOv2~\cite{ruckert2024rocov2}.
Several datasets have introduced volumetric data to support 3D understanding, including CT-RATE~\cite{Ct2rep}, M3D-Data~\cite{M3DLaMed2022}, and MedMD~\cite{RadFM2022}. 
However, functional imaging modalities such as PET/CT remain largely absent from current benchmarks. 
Although datasets like RIDER Lung PET-CT~\cite{muzi2015riderlungpetct}, Lung-PET-CT-Dx~\cite{li2020lungpetctdx}, Head-Neck-PET-CT~\cite{vallieres2017headneckpetct}, and FDG-PET-CT-Lesions \cite{gatidis2022fdgpetctlesions} provide 3D PET/CT scans, they do not include aligned clinical reports, which limits their use for generative modeling and multimodal reasoning.
This gap motivates the need for PET/CT imaging and structured clinical language datasets to enable training and evaluation of VLMs in functional imaging contexts.

To overcome current limitations, we present ViMed-PET, a comprehensive PET/CT image-report dataset composed of two main parts, as summarized in Table~\ref{tab:overview_dataset}.
The first part consists of the original dataset, including paired PET/CT images and their corresponding clinical reports, collected directly from a hospital (Table~\ref{tab:original_data}a).
The second part comprises a series of augmented datasets derived from the original data. These augmentations aim to increase the diversity and richness of the dataset while enabling more fine-grained alignment between the visual and textual modalities (Table~\ref{tab:augmented_data}b).
The following sections detail the structure of ViMed-PET and the methodology used to construct it.

\subsection{Data Description}
The proposed ViMed-PET dataset is collected exclusively from a national tertiary general hospital in Vietnam, one of the country’s largest medical centers. As a high-volume referral institution receiving patients from across all regions, its data reflect broad clinical diversity and ensure high representativeness and reliability.
It consists of 2,757 paired CT-PET volumes (equivalent to 1,567,062 paired CT-PET slices) collected over four years, each accompanied by a corresponding full-length clinical report.
Note that the dataset does not contain complete data for all 12 months of each year, as detailed in the statistics shown in Table~\ref{tab:overview_dataset}. 
Each study includes approximately 250–500 paired CT and PET slices, covering the area from the head to the upper thighs (just above the knees). 
The dataset encompasses various disease cases such as lung cancer, thyroid cancer, and other conditions, representing a broad range of clinical scenarios.
The images are stored in the Digital Imaging and Communications in Medicine (DICOM) format, including pixel data and relevant metadata such as patient age, sex, body weight, radiotracer activity, and other acquisition parameters.
Acquired using GE Discovery 710 PET/CT and GE Discovery STE PET/CT systems, the images provide high-quality data for analysis. 
Furthermore, the PET images have undergone attenuation correction using the corresponding CT data to ensure accurate representation.
Each medical report, stored in DOCX format, corresponds to a single PET/CT study and contains detailed patient information, clinical status, medical history, scanning methods, and physician observations, making the dataset rich in both imaging and clinical information.
All data are obtained under the oversight of an Institutional Ethics Committee (Ethics Approval No. 6184/CN-HDDDBV).
Informed consent for anonymized research use is obtained by the hospital, in accordance with institutional policy and national regulations.

\begin{table}[t]
\caption{The proposed ViMed-PET dataset. 
(a) Statistics of the original data. 
(b) Augmented datasets for training and evaluation various downstream tasks.}
\label{tab:overview_dataset}
\centering
\begin{minipage}[t]{0.66\linewidth}
\label{tab:original_data}
\centering
\vspace{3pt}
{\small(a) Our ViMed-PET dataset {(M: Male, F: Female)}.}
\setlength\tabcolsep{3pt} 
\resizebox{\linewidth}{!}{
\begin{tabular}{l|r|r|c|l|r}
\toprule
\textbf{Year} & \multicolumn{1}{c|}{\textbf{Studies (M, F)}}  & \multicolumn{1}{c|}{\textbf{Age (years)}} & \multicolumn{1}{c|}{\textbf{Height (cm)}} & \multicolumn{1}{c|}{\textbf{Weight (kg)}} & \multicolumn{1}{c}{\textbf{\# Slices}}\\
\midrule
2017 & 215 (137, 78) & 53.55 $\pm$ 15.25 & 160.63 $\pm$ 11.81 & 55.81 $\pm$ 12.02 &  126,766 \\ 
2018 & 462 (308, 154) & 56.77 $\pm$ 13.56 & 161.55 $\pm$ \hspace{2pt} 8.27 & 56.72 $\pm$ \hspace{5pt}9.97 &  270,668 \\
2019 & 339 (227, 112) & 57.35 $\pm$ 12.94 & 161.89 $\pm$ \hspace{2pt} 7.89 & 58.18 $\pm$ \hspace{5pt}9.79 &  200,660 \\
2023 & 1741 (1144, 597)   & 58.69 $\pm$ 13.61 & 161.25 $\pm$ \hspace{2pt} 8.68 & 57.53 $\pm$ 10.10 &  968,968 \\
\midrule
Total & 2757 (1816, 941) & 57.81 $\pm$ 13.73  & 161.33 $\pm$ \hspace{2pt} 8.81 & 57.34 $\pm$ 10.22 & 1,567,062\\
\bottomrule
\end{tabular}
}
\end{minipage}
\hfill
\begin{minipage}[t]{0.33\linewidth}
\centering
\setlength\tabcolsep{3pt} 
\vspace{3pt}
{\small (b) Task-specific augmented dataset.}
\resizebox{\linewidth}{!}{
\begin{tabular}{l|l}
\toprule
\textbf{Subset} & \multicolumn{1}{c}{\textbf{Size}} \\
\midrule
VQA                   & 8,271 conversations \\
Report Generation     & 5,571 reports \\
Study Comparison      & 10,000 pairs \\
Medical Test Set      & 398 lesions \\
\bottomrule
\end{tabular}
\label{tab:augmented_data}
}
\end{minipage}
\end{table}

\subsection{Data Pre-processing}
\begin{figure*}[t]
    \centering
    \includegraphics[width=1.0\textwidth]{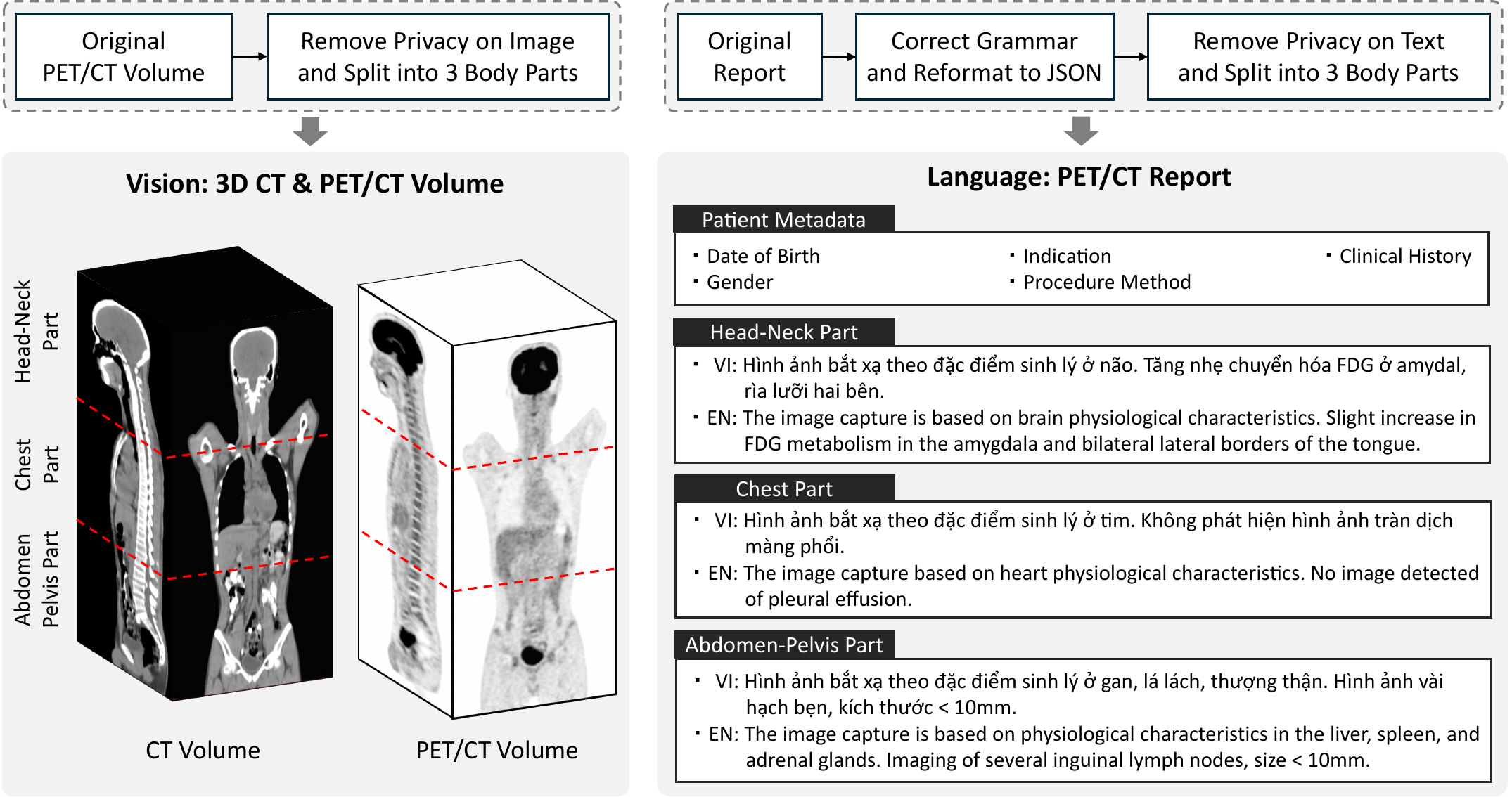}
    \caption{
    An example from our ViMed-PET dataset. The visual input consists of aligned 3D CT and PET/CT volumes, segmented by anatomical regions: head–neck, chest, and abdomen–pelvis. The corresponding report includes patient metadata and structured descriptions for each region in Vietnamese (VI), with English (EN) as the translation.}
    \label{fig:overview_dataset}
\end{figure*}

The pre-processing pipeline consists of several key steps to ensure consistency, accuracy, and usability for model training and evaluation. 
Figure~\ref{fig:overview_dataset} illustrates an example from the ViMed-PET dataset, showing the workflow from raw input to the PET/CT image-report pair.

\textbf{De-identification.} In accordance with privacy regulations, we remove all patient-identifiable information, such as patient name and patient ID, from both the PET/CT images and the associated reports. 
Additionally, to protect confidentiality, we remove details related to doctors and hospitals, including the institution name, the referring physician's name, the names of the physicians reading the study, and the operator's name.

\textbf{Report parsing and standardization.} We structure the reports using a predefined template from the hospital, ensuring consistent formatting. 
A keyword-based search algorithm extracts key information, after which the original DOCX format is converted into a standardized JSON format for easy integration with the corresponding images. 
After automated extraction, we conduct manual checks to correct grammatical or spelling errors, ensuring the integrity of the text data.

\textbf{Body part-based data partition.} 
To expand the dataset and improve the data quality used for training and evaluating the model, we divide each study, comprising a PET/CT volume and its corresponding report, into three anatomically distinct regions: head-neck, chest, and abdomen-pelvis. This approach results in a total of 8,271 paired image-report samples. To preserve continuity across adjacent regions and avoid the loss of critical contextual information at segment boundaries, we introduce a 20-slice overlap between neighboring segments.
The proportional boundaries of each region are adaptively determined based on the patient’s body height, ensuring flexibility across subjects of different statures. The specific ratios for region division follow expert guidelines, with the head–neck typically covering approximately the first 20\% of the body length, the chest starting around 15\% below the last slice of the head–neck region, and the abdomen–pelvis including the remainder from the chest down to the pelvis.
This segmentation strategy increases the number of training samples and significantly improves the alignment between the visual and textual modalities. By localizing image content and clinical descriptions to specific anatomical regions, we enable more precise fine-tuning of models, ultimately enhancing their ability to learn region-specific patterns and improving overall performance.
    

\subsection{Dataset Construction}
\label{subsec:dataset}
We construct four specialized subsets to facilitate various stages of model development and evaluation: (1) Visual Question Answering (VQA), (2) Report generation, (3) Study comparison dataset, and (4) Medical test set. The first three subsets are used to fine-tune the 3D vision encoder and the large language model. 
The final subset is specifically curated to assess the clinical efficacy of the proposed framework in the real-world medical context of a specific task: lung cancer diagnosis.

\textbf{Visual question answering dataset.} 
The VQA dataset aims to fine-tune VLMs by enabling context-aware, multi-turn dialogue about biomedical images for tasks like diagnostic reasoning and clinical decision-making.
The dataset is composed of two parts: single-turn and multi-turn conversations.
First, we created a set of 27,855 image-related questions that prompt descriptive answers. 
These questions were then randomly sampled and paired with each PET/CT image-report pair from the original dataset to generate single-turn VQA samples.
Next, we employ the few-shot prompting strategy to guide the GPT-4o~\cite{hurst2024gpt} in generating multi-turn conversations based on the clinical reports from the original dataset (detailed in the Appendix \ref{appendix:VQA}).
In total, we construct 8,271 multi-turn conversations as part of this dataset.

\textbf{Report generation dataset.} \label{sec:ReportGenData}
To increase the diversity of the dataset, we augment the textual modality. Specifically, we use GPT-4o~\cite{hurst2024gpt} to paraphrase the original clinical reports in the initial dataset. For each original report, we generate one corresponding paraphrased version. To ensure the clinical accuracy of the paraphrased content, a subset of the generated reports is randomly reviewed by medical experts. As a result, starting from the original set of 5,571 paired PET/CT image-report samples, we create an augmented dataset of 5,571 additional PET/CT image-report pairs, which we subsequently use for fine-tuning the model.

\textbf{Study comparison dataset.} \label{sec:StudyComparData}
To further expand the dataset, we introduce a novel augmentation method.
In this approach, instead of aligning a single PET/CT image with its corresponding report, each data instance aligns both the similarity and difference between two PET/CT images with the similarity and difference between their corresponding reports.
Specifically, we construct a study comparison dataset consisting of tuples in the form: $(X_i^u \| X_i^v), \text{Comp} (X_r^u, X_r^v)$, where $(X_i^u \| X_i^v)$ is the concatenation of two PET images from studies $u$ and $v$, and $\text{Comp}(X_r^u, X_r^v)$ is a text description that highlights the comparison between their corresponding reports, including similarities and differences.
The comparison descriptions $\text{Comp}(X_r^u, X_r^v)$ are generated using GPT-4o~\cite{hurst2024gpt} (detailed in the Appendix~\ref{appendix:study-comp}).
This dataset contains a total of 10,000 samples.

\textbf{Medical test set.} 
One of the most challenging aspects of evaluating medical VLMs is quantifying their clinical accuracy. In a generated clinical report, the medical importance and semantic weight of each word vary significantly. For instance, key elements such as lesion type and lesion location carry far more clinical relevance than other details. Therefore, standard NLP evaluation metrics that compare model output with the original report often fail to reflect the true clinical quality of the model's output. To address this challenge, we create a specialized ground-truth dataset that extracts the most clinically significant information from the original reports (refer as \textit{medical-important information}). Such information is then represented in a structured JSON format includes details such as lesion type, lesion location, and key PET/CT metabolism parameters, including SUVmax, FDG metabolism, and the invasiveness of the lesion. The dataset construction follows a two-step process: First, medical experts manually curate a small set of medical-important information from the original reports. Next, we use this curated dataset along with few-shot prompting techniques to guide GPT-4o~\cite{hurst2024gpt} in automatically extracting medical-important information from the full set of clinical reports. To ensure the reliability of the extracted data, all outputs generated by the model are independently verified by two experienced physicians. 
As this study is constrained by available resources, this dataset only focuses on lung cancer cases. In total, we construct a dataset of 80 instances, corresponding to 80 lung cancer patients, covering 398 individual lesions.


\begin{figure*}[t]
    \centering
    \includegraphics[width=1.0\textwidth]{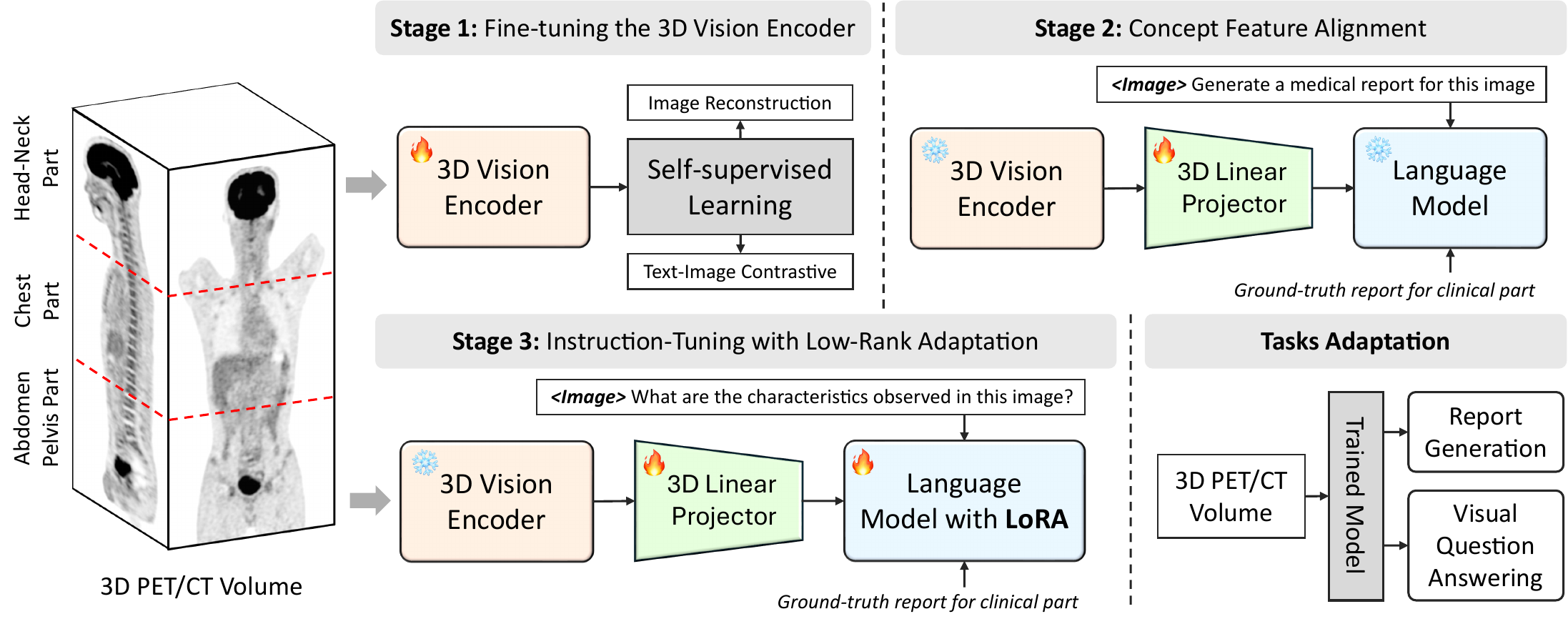}
    \caption{Overview of the fine-tuning pipeline.  {Stage 1:} Fine-tuning the 3D Vision Encoder with PET/CT volumes and associated reports; {Stage 2:} Aligning concept features between the 3D image and textual embeddings; {Stage 3:} Instruction tuning of the complete architecture using Low-Rank Adaptation. 
    Finally, the trained model is used for Report Generation and VQA tasks.}
    \label{fig:framework}
\end{figure*}

%% file: tex/3-benchmarks.tex
In the following, we describe the models used for benchmarking in Section~\ref{sec:model_selection}, and the details of our process to fine-tune these models using ViMed-PET dataset in Section~\ref{sec:finetune}.

\subsection{Model Selection}
\label{sec:model_selection}
A typical VLM consists of two main components: a vision encoder and a text encoder. 
In our framework, we adopt two 3D vision encoders: CT-ViT~\cite{CT-ViT} and Cosmos Tokenizer~\cite{Cosmos}, each selected for their complementary strengths in medical image modeling. 
CT-ViT is chosen as it is the only publicly available vision transformer pretrained on 3D medical imaging data, specifically 3D CT scans. 
In contrast, Cosmos Tokenizer is originally pretrained on general-purpose tasks.
Although not specifically trained on medical images, its architectural design, optimized for handling sequential inputs, makes it naturally compatible with 3D medical imaging, where volumetric scans can be viewed as ordered slices. 
We adapt the Cosmos Tokenizer for 3D PET/CT data by removing the causality-based attention mechanisms that are essential for video modeling but irrelevant for spatially coherent medical volumes. 
For the language component, we utilize Mistral-7B~\cite{jiang2023mistral7b} and LLaMA-2-7B~\cite{touvron2023llama}, language models recommended by the state-of-the-art VLMs as LLaVA-Med~\cite{LLaVAMed2023} and M3D~\cite{M3DLaMed2022}, respectively. 
These models are fine-tuned on biomedical instruction-following datasets, enhancing their ability to interpret complex clinical narratives and engage in nuanced, medically relevant dialogue. 

\subsection{Model Fine-tuning Flow}
\label{sec:finetune}
We employ our curated dataset to fine-tune the baseline models through a structured approach. As shown in Figure~\ref{fig:framework}, our process consists of three stages: (1) Adaptation of the vision encoder to the PET/CT imaging modality, (2) Alignment of visual and textual embedding spaces, and (3) Instruction tuning to optimize the model for downstream multimodal clinical tasks.

\textbf{Stage 1:  3D vision encoder fine-tuning.} 
For the encoder based on CT-ViT~\cite{CT-ViT}, we use a self-supervised learning approach based on text–image contrastive learning, inspired by CLIP~\cite{radford2021learning}.
Here, the model is trained to align visual representations of PET images with textual embeddings derived from their corresponding clinical reports. 
This cross-modal supervision enables the encoder to learn semantically meaningful and clinically relevant visual features.
In contrast, for the encoder based on Cosmos Tokenizer~\cite{Cosmos}, we employ an image reconstruction objective.
Specifically, a decoder within the Cosmos Tokenizer architecture is used to reconstruct the original PET/CT volume from the latent representations produced by the encoder.
The model is trained with a reconstruction loss, which encourages it to capture the underlying structural and semantic features of the 3D medical images.

\textbf{Stage 2: Concept feature alignment.}
In this stage, we fine-tune a linear projection layer that maps visual feature representations into the embedding space of the text encoder. 
We use single-turn image-text pairs from our VQA dataset, where each sample includes a PET/CT image and an instruction-based question. 
Examples include prompts such as ``<image> What are the main findings in this medical image?'' or ``<image> Please write a detailed medical report for this image.''.
The model is trained to generate the original textual response based on the given image and prompt.
During this process, we freeze the weights of the visual encoder and the language model, allowing updates only to the linear projector. 
This design ensures efficient and stable alignment between visual and textual embeddings while minimizing overfitting and preserving pre-trained representations.
%

\textbf{Stage 3: Instruction-tuning with Low-Rank Adaptation.} 
In the final stage, we fine-tune the VLM using our VQA dataset, which includes single-turn question–answer pairs and multi-turn conversational interactions.
During this phase, the parameters of the image encoder are kept frozen to preserve previously learned visual representations.
We update only the parameters of the visual–text alignment projector and the language model, applying Low-Rank Adaptation (LoRA)~\cite{hu2022lora} for efficient fine-tuning.
This stage strengthens the capacity of the model to interpret and respond to a broad spectrum of medical queries by integrating visual and textual modalities. 
The inclusion of simple and complex dialogue formats further improves its robustness in biomedical VQA tasks.
%

\subsection{Training and Evaluation}
We integrate two vision encoders (CT-ViT and a customized Cosmos Tokenizer) with two text encoders (Mistral-7B and LLaMA-2-7B) to construct four VLMs. 
We evaluate the performance of these models on two tasks: medical report generation and VQA.
The dataset is divided into three subsets: a training set with 5,571 image-report pairs, a validation set with 975 pairs, and a test set with 1,725 pairs. Each subset contains samples from all four years of our collection to reduce temporal bias.
To assess the impact of data augmentation, we fine-tune our framework under three configurations: (1) {O}: using only the Original dataset, (2) {O-G}: using the Original dataset and the report Generation dataset, (3) {O-G-C}: using the Original dataset, the report Generation dataset, and the study Comparison dataset. 
In addition, we benchmark our models against baselines, including LLaVA-Med~\cite{LLaVAMed2023}, M3D~\cite{M3DLaMed2022}, RadFM~\cite{RadFM2022}, and GPT-4o~\cite{hurst2024gpt}.
Further details regarding model architectures, training configurations, and optimization settings are provided in Appendix~\ref{appendix:train_model_config}.

\subsection{Evaluation Metrics} 
\textbf{Common natural language processing metrics.} 
We evaluate model performance using standard Natural Language Processing (NLP) metrics, including BLEU-4~\cite{BLEU}, ROUGE~\cite{ROUGE}, and BERTScore~\cite{BertScore}.
BLEU-4 measures 4-gram precision in generated text, ROUGE-1 captures unigram recall for summarization tasks, and ROUGE-L evaluates fluency based on the longest common subsequence.
BERTScore assesses semantic similarity between generated and reference texts by leveraging contextual embeddings from a pre-trained BERT model.

\textbf{Proposed clinical metrics.} Conventional NLP metrics are insufficient for evaluating the clinical accuracy of generated text, particularly in terms of medical relevance. 
For example, in auto-generated clinical reports, details such as tumor location or FDG uptake carry significantly more diagnostic weight than general descriptions.
To address this limitation, we introduce targeted evaluation metrics that focus specifically on clinically meaningful content.
We use our expert-curated Medical Test Set (see Section~\ref{subsec:dataset}) to evaluate model performance on the report generation task, emphasizing the extraction of key attributes including lesion Type, lesion Position, and FDG uptake.
Specifically, for each generated report, we use GPT-4o~\cite{hurst2024gpt} to extract structured content aligned with these three clinical attributes. The extracted information is converted into categorical variables (detailed in Appendix~\ref{appendix:clinical_eval}) and verified by medical professionals.
We then compute F1-scores by comparing these model-generated outputs with the ground truth labels in our test set.
We report four clinical evaluation metrics: F1-T, based on lesion Type; F1-TP, based on Type and Position; F1-TF, incorporating Type and FDG uptake; and F1-TPF, which evaluates all three aspects of Type, Position, and FDG uptake.

%% file: tex/4-evaluation_results.tex
\label{sec:evaluation}
This section presents the experimental results assessing the performance of the VLMs fine-tuned on our ViMed-PET dataset, with respect to two key tasks: clinical report generation (Table \ref{tab:dataset_augmentation_comparison}) and medical VQA (Table \ref{tab:medical}). 
Additional benchmarks and evaluation results are provided in Appendix~\ref{appendix:additional_res}.

\begin{table*}[t]
\centering
\caption{Performance on report generation task. We define training configurations as: \textbf{O}-\textbf{O}riginal dataset, \textbf{G}-Report \textbf{G}enerate dataset, \textbf{C}-Study \textbf{C}omparison dataset. R-1 and R-L denote ROUGE-1 and ROUGE-L scores. \textbf{$\uparrow$} means higher values are better. The best and second-best results are emphasized using \best{bold} and \second{underline}, respectively.
\textit{*GPT-4o is evaluated under few-shot prompting.}
} 
\label{tab:dataset_augmentation_comparison}
\resizebox{\textwidth}{!}{
\begin{tabular}{l|l|l|ccc|cccc|cccc}
\toprule
& \multicolumn{2}{c|}{\textbf{Model}} & \multicolumn{3}{c|}{\textbf{Settings}} & \multicolumn{4}{c|}{\textbf{NLP Metrics $\uparrow$}} & \multicolumn{4}{c}{\textbf{Clinical F1-Score (\%) $\uparrow$}} \\
\cmidrule(lr){2-3} \cmidrule(lr){4-6} \cmidrule(lr){7-10} \cmidrule(lr){11-14}
 & Vision & Language & O & G & C & BLEU-4 & R-1 & R-L & BERT & F1-T & F1-TP & F1-TF & F1-TPF \\
\midrule
\multirow{4}{*}{\rotatebox[origin=c]{90}{\textbf{Baseline}}} & \multicolumn{2}{c|}{LLaVA-Med~\cite{LLaVAMed2023}}  &   &  --  &    & \hspace{2pt} 0.01 & 50.08 & 27.89 & 64.63 & -- & -- & -- & -- \\
& \multicolumn{2}{c|}{M3D~\cite{M3DLaMed2022}}  &   &   -- &    & \hspace{2pt} 0.04 & 41.01 & 23.53 & 67.21 & -- & -- & -- & -- \\
& \multicolumn{2}{c|}{RadFM~\cite{RadFM2022}}  &   &  --  &    & \hspace{2pt} 0.06 & 54.23 & 28.33 & 69.49 & -- & -- & -- & -- \\
& \multicolumn{2}{c|}{GPT-4o*~\cite{hurst2024gpt}} &   &  --  &    & 31.12 & 67.96 & 52.76 & 81.09 & 24.21 & 13.62 & 20.57 & 7.87 \\
\midrule

\multirow{12}{*}{\rotatebox[origin=c]{90}{\textbf{Fine-tuned}}} & \multirowcell{6}[-2pt][l]{CT-ViT} & & \checkmark  & &       & 53.30 & 77.79 & 68.60 & 88.35 & 43.49 & 24.97 & 29.80 & 18.26  \\
          & & Mistral-7B & \checkmark  & \checkmark &         & \second{58.07} & \best{80.11} & 72.74 & 89.98 & \second{51.11}  & \second{30.66}  & \best{37.02} & \best{22.65}  \\
          & & & \checkmark  & \checkmark & \checkmark    & 58.05 & 80.08 & 72.70 & 89.92 & \best{51.96} & 30.17 & \second{35.47} & 21.23\\
          \cmidrule(lr){3-14}
  & &    & \checkmark  & &          & 57.91 & 79.87 & 72.89 & 89.88 & 48.28 & 29.44 & 33.42 & 20.42 \\
           & & LLaMA-2-7B & \checkmark  & \checkmark &         & 56.15 & 79.09 & 71.39 & 89.20 & 46.24 & 24.15 & 32.40 & 17.38 \\
           & & & \checkmark  & \checkmark & \checkmark    & 56.05 & 78.95 & 71.43 & 89.12 & 47.97 & 29.17 & 33.66 & 20.76\\
           
\cmidrule(lr){2-14}

& \multirowcell{6}[-2pt][l]{Cosmos \\ Tokenizer} &  & \checkmark  & &          & 53.66 & 78.04 & 69.93 & 88.69 & 46.29 & 30.00 & 31.14 & 19.71 \\
           & & Mistral-7B & \checkmark  & \checkmark &         & 55.80 & 78.93 & 71.68 & 89.43 & 46.51  & 25.68 & 31.10 & 17.40 \\
           & & & \checkmark  & \checkmark &  \checkmark   & \best{58.87} & \second{80.10} & \best{73.91} & \best{90.55} & 47.93   & 22.78  & 32.84 & 15.68  \\ \cmidrule(lr){3-14}
  & &  & \checkmark  &  &          & 57.59 & 79.26 & \best{73.91} & \second{90.35} & 45.05 & 18.21 & 32.27 & 14.06 \\
           & & LLaMA-2-7B & \checkmark  & \checkmark &         & 57.98 & 79.83 & \second{73.49} & 90.04 & 45.59 & 27.13 & 30.21 & 17.62 \\
           & & & \checkmark  & \checkmark & \checkmark    & 57.14 & 79.37 & 73.13 & 90.01 & 49.73  & \best{31.42} & 33.61 & \second{22.13} \\

\bottomrule
\end{tabular}
}
\end{table*}

\subsection{PET/CT Report Generation Task} 
\textbf{Comparison with existing baselines.} 
Our results first demonstrate that fine-tuning VLMs on our proposed ViMed-PET dataset leads to substantial performance improvements across both standard NLP metrics and clinically specific evaluation metrics. 
For instance, when LLaMA-2-7B is paired with either CT-ViT or our customized Cosmos Tokenizer and fine-tuned on ViMed-PET, it significantly outperforms the pretrained LLaMA-2-7B model used in M3D across all key metrics (i.e., BLEU-4, ROUGE-1, ROUGE-L, and BERT score). 
Notably, the BLEU-4 score improves significantly over baseline medical LLMs following fine-tuning, reflecting a dramatic enhancement in generation quality.
Compared to GPT-4o, which is evaluated under few-shot prompting, models fine-tuned with ViMed-PET also yield substantial performance gains.
Specifically, BLEU-4, ROUGE-1, ROUGE-L, and BERT score increase by up to 89.17\%, 17.88\%, 40.09\%, and 11.67\%, respectively. 
Furthermore, clinical metrics, including F1-T, F1-TP, F1-TF and F1-TPF, also exhibit notable improvements, increasing by more than 1.8 times, underscoring the clinical relevance and robustness of the generated reports.
These findings highlight the effectiveness of fine-tuning with ViMed-PET in enhancing the performance of VLMs for clinical report generation.

\textbf{Comparison between LLMs.}
Comparing the performance of the two LLMs using common NLP metrics, we observe that when fine-tuning is performed solely on the original dataset (setting O), LLaMA2-7B outperforms Mistral-7B.
This can be attributed to the fact that the M3D backbone used in LLaMA2-7B has been pretrained on 3D medical imaging data, allowing it to better capture spatial features and align 3D PET/CT representations with textual descriptions. 
In contrast, Mistral-7B is pretrained on 2D image–text data, which limits its capacity to model 3D spatial context in low-data settings.
However, when the training dataset is expanded to include both the original and augmented data (settings O-G and O-G-C), Mistral-7B demonstrates superior performance, likely due to its more efficient architecture compared to LLaMA2-13B. 
This finding is consistent with results reported in the Mistral-7B paper~\cite{jiang2023mistral7b}, which highlights the robustness of the model and scalability in large-scale learning settings.
For clinical evaluation metrics, Mistral-7B outperforms LLaMA2-7B when paired with the CT-ViT encoder. However, when combined with the Cosmos Tokenizer, LLaMA2-7B shows a slight advantage over Mistral-7B.

\textbf{Comparison between vision encoders.} 
The results show that CT-ViT and Cosmos Tokenizer achieve comparable performance on standard NLP metrics across various settings, including integration with different LLMs and the use of augmented data.
However, CT-ViT consistently outperforms Cosmos Tokenizer on clinical metrics across all four evaluation criteria and training configurations. 
This indicates that CT-ViT, which is specifically designed and pretrained on 3D medical imaging data, is more effective in improving the clinical accuracy of VLMs than Cosmos Tokenizer, which is pretrained on general-purpose tasks.

\subsection{PET/CT VQA Task}
\begin{table}[t]
\caption{Performance on the VQA task under the O-G-C training setting.
R-1 and R-L denote ROUGE-1 and ROUGE-L scores.
{\textbf{$\uparrow$} means higher values are better. The best and second-best results are emphasized using \best{bold} and \second{underline}, respectively.} \textit{*GPT-4o is evaluated under few-shot prompting.}}
\label{tab:medical}
\centering
\resizebox{0.6\linewidth}{!}
{
\begin{tabular}{l|l|cccc}
\toprule
\multicolumn{2}{c|}{\textbf{Model}} &  \multicolumn{4}{c}{\textbf{NLP Metrics $\uparrow$}} \\ 
\cmidrule(lr){1-6}
Vision & Language & {BLEU-4} & {R-1} & {R-L} & {BERT} \\
\midrule
\multicolumn{2}{c|}{GPT-4o*~\cite{hurst2024gpt}} & \hspace{2pt} 3.01 & 49.35  & 30.09 & 71.92 \\
\midrule
 \multirowcell{2}[-2pt][l]{CT-ViT} & Mistral-7B & \second{31.14} & \best{65.61} & \best{51.22} & \best{82.50} \\
\cmidrule(lr){2-6}
& LLaMA-2-7B & \best{31.36} & 59.14 & 48.00 & 76.72 \\
\midrule
 \multirowcell{2}[-2pt][l]{Cosmos \\ Tokenizer} & Mistral-7B & 28.09 & 62.92 & 48.37 & 79.25 \\
\cmidrule(lr){2-6}
& LLaMA-2-7B & 28.40 & \second{63.29} & \second{48.76} & \second{79.35} \\
\bottomrule
\end{tabular}
}
\end{table}


Table~\ref{tab:medical} shows the results on the VQA task under the O-G-C training setting. 
Fine-tuning with our ViMed-PET dataset significantly improves performance across all evaluation metrics compared to the baseline GPT-4o model.
Specifically, the best-performing fine-tuned model, CT-ViT paired with Mistral-7B, achieves substantial gains, outperforming GPT-4o by factors of $10.3\times$, $1.3\times$, $1.7\times$, and $1.1\times$ on BLEU-4, ROUGE-1, ROUGE-L, and BERT score, respectively.
When comparing different combinations of vision and language encoders, the results on the VQA task follow a pattern similar to those in the report generation task.
The combination of CT-ViT and Mistral-7B consistently delivers the highest performance, followed by Cosmos Tokenizer paired with LLaMA-2-7B.
A comprehensive analysis of the VQA task is provided in the Appendix~\ref{appendix:additional_res}.

%% file: tex/5-conclusion.tex
In this study, we introduced ViMed-PET, a high-quality dataset comprising 2,757 paired whole-body PET/CT volumes and 2,757 Vietnamese clinical reports, covering a wide range of patient cases.
We also developed a clinically validated lung cancer test set to support meaningful evaluation beyond conventional NLP metrics.
Additionally, we proposed a data augmentation strategy that enhances both visual and textual inputs to improve the fine-tuning of medical vision-language models.
Models fine-tuned on ViMed-PET demonstrated substantial gains, achieving improvements in both standard NLP metrics and clinical evaluation scores compared to pretrained baselines.

\textbf{Limitations and Societal Impacts.} We acknowledge that clinical reports often follow standardized formats, which can limit output diversity. Clinical results with F1 scores around 50\% highlight the challenges in modeling PET/CT data and performing complex medical reasoning. 
Although the proposed ViMed-PET dataset contains PET and CT volumes, our benchmark focuses exclusively on PET/CT imaging. 
This decision is based on the observation that report content is primarily driven by PET information, with minimal reference to CT anatomical details. 
In future work, we plan to extend our approach to better incorporate CT data and further improve the accuracy of VLMs. 
This study provides a foundation for such enhancements. 
We believe ViMed-PET serves as a valuable resource for advancing medical vision-language modeling in the low-resource Vietnamese language and the underexplored PET/CT domain, supporting more equitable AI development in healthcare.

\section*{Acknowledgments}

This research was supported by the NVIDIA Academic Grant Program. We gratefully acknowledge NVIDIA for providing the computational resources used in this study, where all model training and evaluation were conducted on NVIDIA A100 GPUs. We also thank NVIDIA for publicly releasing the Cosmos Tokenizer, which served as one of the 3D vision encoders in our benchmark.

%% file: tex/2-related_work.tex

Recent advances in  Vision-Language Models (VLMs) have opened new possibilities for medical image analysis by enabling multimodal reasoning across visual and textual inputs. 
In the medical domain, VLMs are broadly categorized into two types: CLIP-based and Large-Language-Model-based (LLM-based) models.
CLIP-based models, such as MedCLIP~\cite{MedCLIP2021}, PLIP~\cite{PLIP2021}, and BiomedCLIP~\cite{BiomedCLIP2021}, leverage contrastive learning to align images with textual descriptions, performing well on classification and retrieval tasks.
However, their lack of generative capability limits their use in applications such as report generation.
In contrast, LLM-based models, including M3D-LaMed~\cite{M3DLaMed2022}, CT2Rep~\cite{Ct2rep}, Merlin~\cite{Merlin}, and RadFM~\cite{RadFM2022}, combine image encoders with language models to support complex reasoning and text generation.

Despite recent advances, most existing VLMs are trained primarily on 2D medical images (e.g., X-rays, Dermatology, Pathology), with the models such as XrayGPT~\cite{XrayGPT}, ELIXR~\cite{ELIXR}, and CheXagent~\cite{chen2024chexagent}. 
This focus limits their capacity to process 3D imaging modalities like PET/CT, which require spatial and intensity-aware reasoning across volumetric data.
In addition, most VLMs are developed for English, with limited support for other languages due to a lack of multilingual annotated datasets. 
Recent models like M3D-LaMed~\cite{M3DLaMed2022} and RadFM~\cite{RadFM2022} introduce architectures capable of handling 3D inputs, improving performance across imaging modalities. 
For multilingual contexts, Qilin-Med-VL~\cite{liu2023qilin} and HuatuoGPT-Vision~\cite{zhang2023huatuogpt} show potential in Chinese and bilingual applications. 
However, these VLMs perform poorly on PET/CT imaging, often confusing it with MRI or SPECT and failing to produce accurate, medically grounded outputs.
However, these efforts have yet to address the needs of low-resource languages such as Vietnamese, where both medical imaging and language data remain scarce.

%% file: tex/6-appendix.tex
\subsection{Visual Question Answering Dataset}\label{appendix:VQA}
\begin{figure}[ht]
    \centering
    \includegraphics[width=1.0\textwidth]{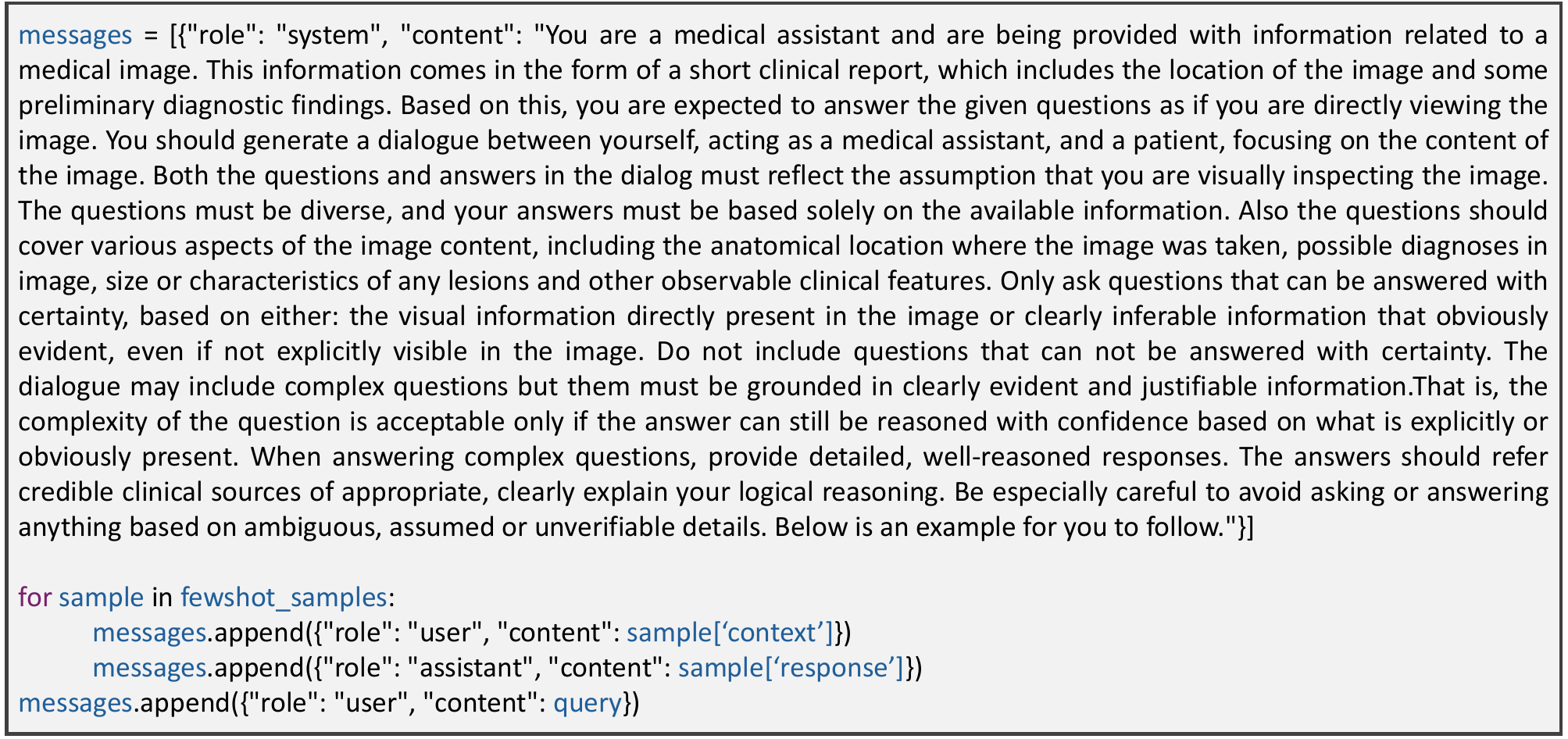}
    \caption{Message used to prompt GPT-4o to generate our medical VQA conversations. Manually curated few-shot examples are included in the prompt, where each example consists of an input \texttt{sample[`context']} and an output \texttt{sample[`response']}. See Figure~\ref{fig:fewshot_ex_vqa_data_gen} for a sample few-shot example.}
    \label{fig:prompt_message_vqa_data_gen}
\end{figure}
\begin{figure}[t]
    \centering
    \includegraphics[width=1.00\textwidth]{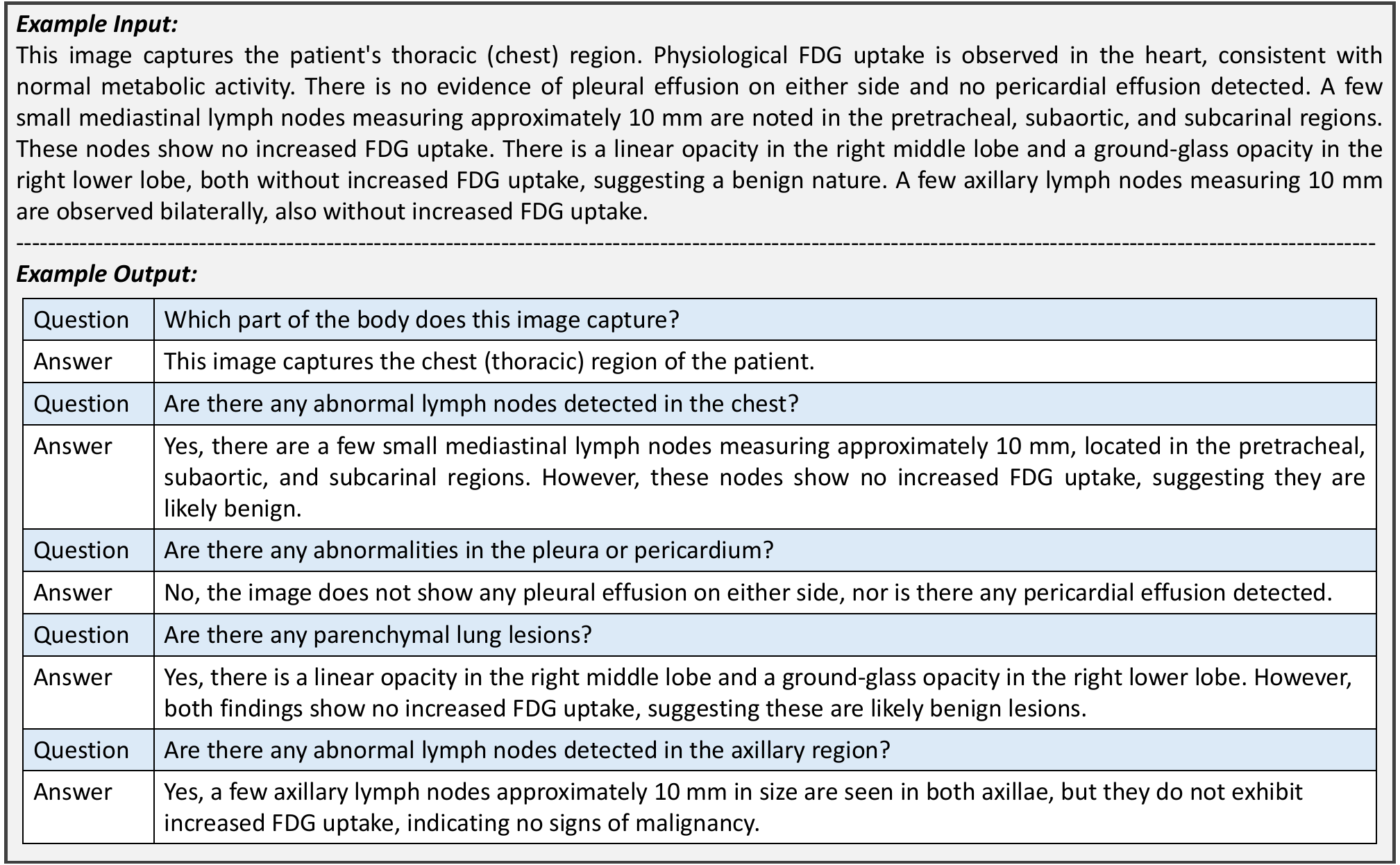}
    \caption{Few-shot examples included in our prompt to construct the VQA conversation dataset.}
    \label{fig:fewshot_ex_vqa_data_gen}
\end{figure}

{
To construct the Visual Question Answering (VQA) dataset, we follow the methodology introduced by LLaVA~\cite{li2024llava}, adopting two types of response formats: \textbf{detailed description} and \textbf{conversation}.
For the detailed description format, we directly use the original clinical report as the textual response corresponding to the input image.
For the conversational format, we employ instruction-based prompting in conjunction with few-shot prompting. 
Specifically, GPT-4o~\cite{hurst2024gpt} is guided to generate coherent question–answer pairs by providing clear instructions that constrain the generated content to align with the given report.
Additionally, we include several few-shot examples to illustrate the desired structure and tone, enhancing the ability of the model to produce contextually appropriate outputs.
The full prompting details are shown in Figures~\ref{fig:prompt_message_vqa_data_gen} and ~\ref{fig:fewshot_ex_vqa_data_gen}.
Note that while the figures are translated into English for illustration purposes, the actual prompts and data are in Vietnamese.
}

\subsection{Study Comparison Dataset}
\label{appendix:study-comp}

\begin{figure*}[h]
    \centering
    \includegraphics[width=\textwidth]{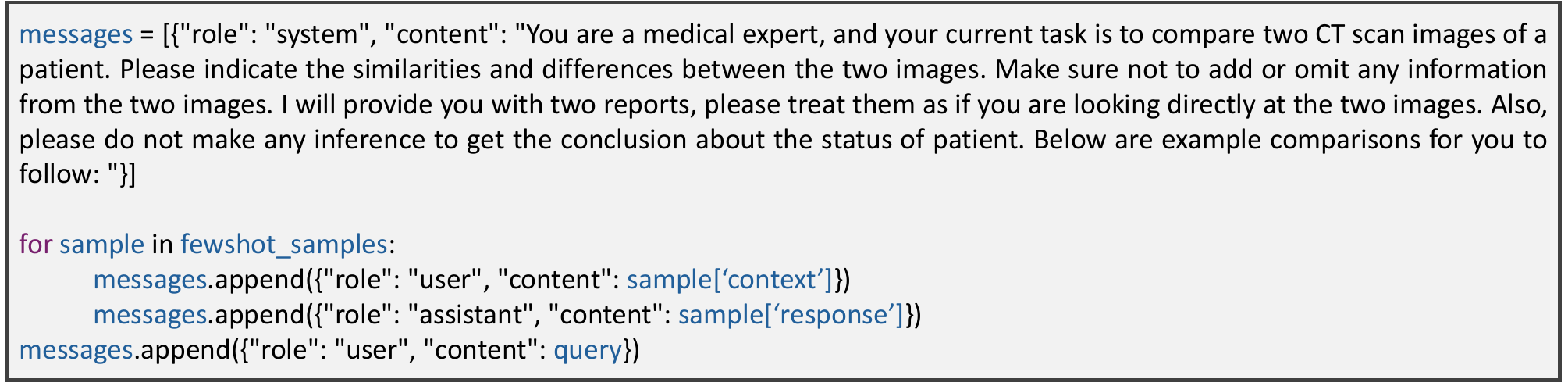}
    \caption{System message used to prompt GPT-4o for generating the study comparison dataset. The prompt includes few-shot examples, where each example consists of an input \texttt{sample[`content']} (a pair of reports to compare) and an output \texttt{sample[`response']} that describes the similarities and differences between the two reports. See Figure~\ref{fig:fewshotB2} for a sample few-shot example.}
    \label{fig:prompt_message}
\end{figure*}

\begin{figure*}[h]
    \centering
    \includegraphics[width = 1.0\textwidth]{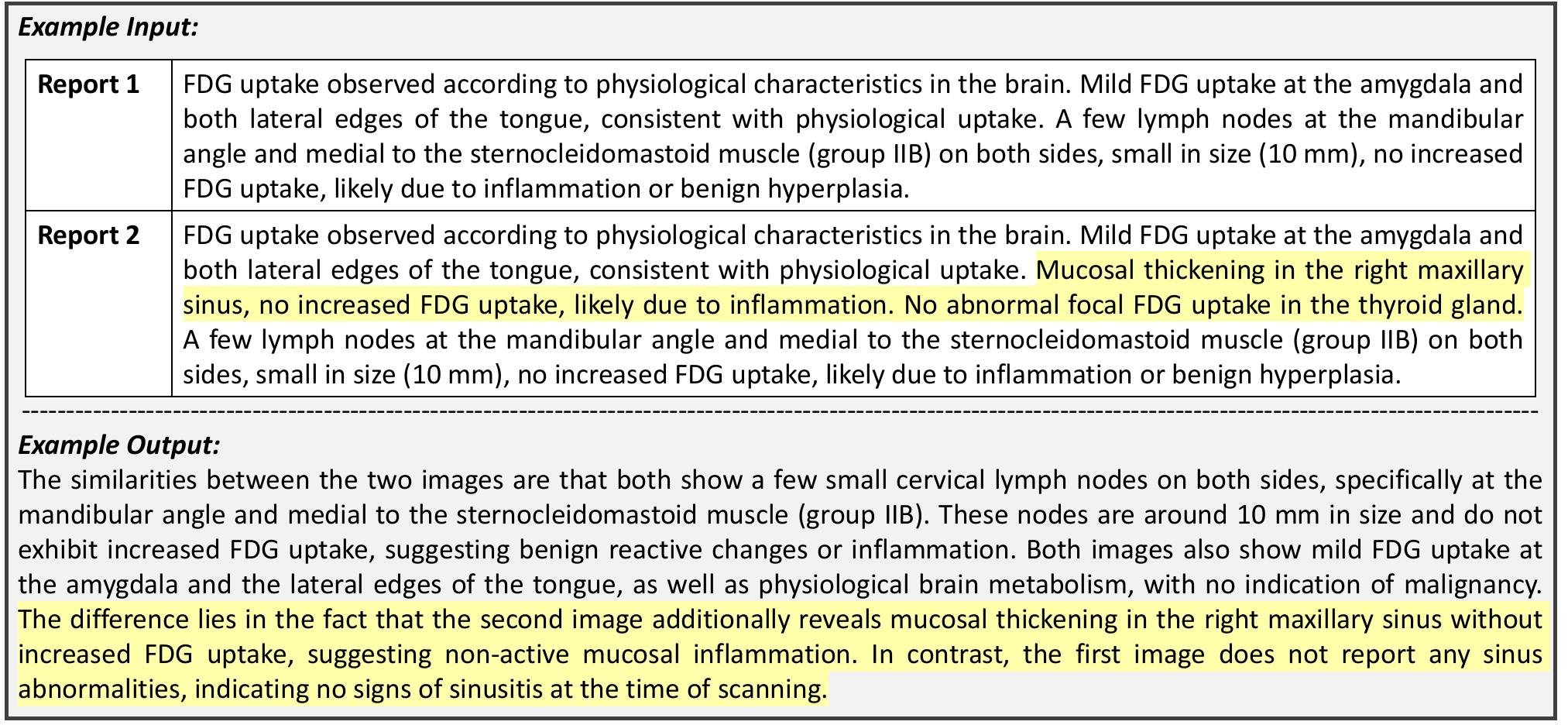}
    \caption{Few-shot example used in our prompt for generating the study comparison dataset. The yellow highlights indicate the differences between the two reports.}
    \label{fig:fewshotB2}
\end{figure*}

{
To construct the study comparison dataset, we adopt a few-shot prompting approach using GPT-4o~\cite{hurst2024gpt}. 
Few-shot examples are created by randomly sampling three pairs of medical reports, each annotated by domain experts to highlight similarities and differences between the reports. 
To ensure meaningful comparisons, all report pairs are selected from the same anatomical region. 
The full prompting setup is illustrated in Figures~\ref{fig:prompt_message} and~\ref{fig:fewshotB2}. 
Note that while the figures are translated into English for illustration purposes, the actual prompts and data are in Vietnamese.
}

\newpage
\subsection{Training and Model Configurations}\label{appendix:train_model_config}
\subsubsection{Fine-tuning Vision Encoders}
{
We select CT-ViT~\cite{CT-ViT} and Cosmos Tokenizer~\cite{Cosmos} as the vision encoders for our VLMs, as they are well-suited for processing 3D volumetric inputs with depths of up to 200 slices and have been pre-trained on large-scale datasets in prior work. Details on model selection are discussed in Section~3.1.
}

{
\noindent \textbf{CT-ViT.} We employ a specialized ViT model, CT-ViT~\cite{CT-ViT}, as the vision encoder in our VLMs. 
CT-ViT is designed to effectively process 3D chest CT volumes and is pre-trained on a large-scale medical dataset comprising 25,701 non-contrast 3D chest CT volumes from 21,314 unique patients. 
These volumes vary in resolution and contain between 100 and 600 axial slices. 
To align visual and textual modalities, we adopt a CLIP-based~\cite{radford2021learning} training approach. 
The model is fine-tuned for up to 30 epochs using the AdamW optimizer~\cite{loshchilov2017decoupled}, with a learning rate of $1.25 \times 10^{-6}$ and a batch size of 8 per GPU across four NVIDIA A100 GPUs (80 GB each). 
Early stopping is applied based on the convergence of training loss, ensuring efficient optimization.
For the text modality, we integrate PhoBERT~\cite{nguyen2020phobert}, a state-of-the-art Vietnamese language model pre-trained on a large-scale Vietnamese corpus. 
PhoBERT has demonstrated superior performance over multilingual models such as XLM-R across several Vietnamese natural language processing (NLP) tasks, including part-of-speech tagging, dependency parsing, named entity recognition, and natural language inference. 
Its linguistic compatibility with clinical texts in our dataset enables effective semantic representation and understanding.
}

{
\noindent \textbf{Cosmos Tokenizer.}
We leverage the architecture of the Cosmos Tokenizer~\cite{Cosmos}, originally designed for image and video reconstruction tasks. 
To adapt it for 3D PET/CT imaging, we remove causality-based attention mechanisms, which are essential for modeling temporal dependencies in video but unnecessary for spatially coherent volumetric medical scans. 
This modification allows us to retain the benefits of pre-trained weights while enabling effective processing of 3D medical data.
We fine-tune the customized Cosmos Tokenizer using a single-phase reconstruction approach. 
The total loss function $\mathcal{L}_{\text{total}}$ combines two terms: an $L_1$ reconstruction loss $\mathcal{L}_1$ and an inverted Structural Similarity Index Measure (SSIM) loss $\mathcal{L}_{\text{iSSIM}}$, defined as:
\begin{equation}
    \mathcal{L}_{\text{total}} = \mathcal{L}_1 + \lambda \mathcal{L}_{\text{iSSIM}} = \| \hat{x}_{0:T} - x_{0:T} \|_1 + \lambda \big(1 - \text{SSIM}(\hat{x}_{0:T}, x_{0:T}) \big)
\end{equation}
where $\hat{x}_{0:T}$ is the reconstructed volume, $x_{0:T}$ is the ground-truth volume, and $\lambda$ is the trade-off coefficient (set to $1 \times 10^{-2}$ across all experiments).
Training is performed for up to 20 epochs using a Cosine Annealing Scheduler~\cite{loshchilov2016sgdr}, with an initial learning rate of $1 \times 10^{-4}$ and a batch size of 8 across four NVIDIA A100 GPUs (80\,GB each). 
Since the Cosmos Tokenizer requires a fixed number of input frames, we standardize all PET/CT volumes to 120 slices. This value is chosen based on the distribution in our dataset, and we apply zero-padding or linear interpolation to achieve this fixed size.
}

The original Cosmos Tokenizer uses causal attention for video tasks, modeling forward-only temporal relationships. 
However, 3D PET volumes have bidirectional spatial relationships, making causal attention less appropriate. 
To assess this, we compared both settings (with vs.\ without causal masking) using \texttt{LLaMA-2-7B} on the ViMed-PET dataset across two tasks: Report Generation and Visual Question Answering (VQA). 
As shown in Table~\ref{tab:causal_attention}, removing causal attention yields significantly better performance across both tasks.

\begin{table*}[t]
\centering
\caption{Performance comparison of causal attention settings on ViMed-PET across two tasks: Report Generation and Visual Question Answering (VQA). R-1 and R-L denote ROUGE-1 and ROUGE-L scores. $\uparrow$ indicates higher is better.}
\label{tab:causal_attention}
\resizebox{\textwidth}{!}{
\begin{tabular}{l|l|c|cccc|cccc}
\toprule
\multicolumn{2}{c|}{\textbf{Model}} & \multirowcell{2}[1pt][c]{\makecell[c]{\textbf{Causal}\\\textbf{Attention}}} & \multicolumn{4}{c|}{\textbf{Report Generation}} & \multicolumn{4}{c}{\textbf{VQA}} \\
\cmidrule(lr){1-2}  \cmidrule(lr){4-7} \cmidrule(lr){8-11}
\textbf{Vision} & \textbf{Language} &  & BLEU-4$\uparrow$ & R-1$\uparrow$ & R-L$\uparrow$ & BERT$\uparrow$ & BLEU-4$\uparrow$ & R-1$\uparrow$ & R-L$\uparrow$ & BERT$\uparrow$ \\
\midrule
\multirow{2}{*}{\makecell[l]{Cosmos\\Tokenizer}}
& \multirow{2}{*}{LLaMA-2-7B} 
& \checkmark & 54.99 & 77.82 & 68.88 & 87.86 & 19.87 & 55.60 & 41.10 & 76.49 \\
&  & $\times$ & 57.59 & 79.26 & 73.91 & 90.35 & 28.40 & 63.29 & 48.76 & 79.35 \\
\bottomrule
\end{tabular}}
\end{table*}

{
\subsubsection{Fine-tuning VLMs}
After fine-tuning the vision encoders, we integrate each with two language models derived from state-of-the-art medical multimodal foundation models: LLaMA-2-7B from M3D~\cite{M3DLaMed2022} and Mistral-7B from LLaVA-Med~\cite{LLaVAMed2023}. 
The integration is facilitated by a linear projection layer that aligns the visual and textual embedding spaces.
}

{
\noindent\textbf{Conceptual Alignment.} 
We use single-turn data composed of prompts such as ``<image> What are the main findings in this medical image?'' and ``<image> Please write a detailed medical report for this image.'', paired with the corresponding medical report as the target output. 
During training, the weights of both the LLM and the vision encoder are frozen, allowing updates only to the linear projection layer.
Training is conducted using a batch size of 16 per GPU across 4 A100 GPUs (80 GB each), with gradient accumulation over 4 steps. 
We employ the AdamW optimizer~\cite{loshchilov2017decoupled} with a warmup ratio of 0.03 and an initial learning rate of $2 \times 10^{-3}$, followed by a Cosine Annealing Scheduler~\cite{loshchilov2016sgdr}. 
Training runs for up to 20 epochs, and the checkpoint with the lowest validation loss is selected for evaluation.
}

{
\noindent\textbf{LoRA Fine-tuning.} 
We employ both single-turn and multi-turn conversational data to continue fine-tuning the linear projector and to update the LLM using the Low-Rank Adaptation (LoRA)~\cite{hu2022lora} method. 
This method efficiently adapts the pretrained LLM by injecting trainable low-rank matrices into selected linear layers, substantially reducing the number of trainable parameters and computational overhead. The LoRA configuration is set as follows: rank ($r$) = 64, scaling factor ($\alpha$) = 16, and dropout rate = 0.05. The task type is defined as \texttt{CAUSAL\_LM}, aligning with the LLM's causal language modeling objective.
Training is conducted with a batch size of 8 per GPU across 4 NVIDIA A100 GPUs (80 GB each), using gradient accumulation over 4 steps. 
We use the AdamW optimizer~\cite{loshchilov2017decoupled} with a warmup ratio of 0.03 and an initial learning rate of $2 \times 10^{-5}$, followed by a Cosine Annealing Scheduler~\cite{loshchilov2016sgdr}. 
Training is performed for 20 epochs, and the checkpoint with the lowest validation loss is selected for evaluation.
}

{
\subsubsection{Fine-tuning Resources}
We report the training time and GPU memory consumption for fine-tuning VLMs across different stages, using a setup of four NVIDIA A100 GPUs with 80 GB memory each, as summarized in Table~\ref{tab:train_infer_resources}. 
The GPU memory values in the table reflect the peak consumption observed across all four GPUs. 
All measurements were recorded under a consistent software environment: Python 3.8.20, CUDA nvcc 12.8.61, Accelerate 1.0.1, DeepSpeed 0.16.2, PyTorch 2.1.0, Transformers 4.46.3, and PEFT 0.4.0.
Our results show that VLMs utilizing the Cosmos Tokenizer as the vision encoder are more efficient in both training time and memory usage compared to those based on the CT-ViT architecture. 
This suggests that the architectural design of the Cosmos Tokenizer offers a more resource-efficient training process, which is particularly advantageous in large-scale or resource-constrained settings.
}

\begin{table*}[h]
\centering
\caption{Training resource consumption of VLMs on the Original dataset. Memory (Mem) values indicate the peak GPU memory usage (in GB) across four A100 GPUs.}
\label{tab:train_infer_resources}
\resizebox{0.85\textwidth}{!}{
\begin{tabular}{l|l|l|cc|cc}
\toprule
& \multicolumn{2}{c|}{\multirowcell{2}[-2pt][l]{\hspace{23pt} \textbf{Model}}} & \multicolumn{4}{c}{\textbf{Computational Resources}} \\
\cmidrule(lr){4-7}
 & \multicolumn{2}{c|}{} & \multicolumn{2}{c|}{Concept Alignment} & \multicolumn{2}{c}{LoRA Fine-tuning} \\
\cmidrule(lr){2-3} \cmidrule(lr){4-7}
& Vision & Language & Time (Hours) & Mem (GB) & Time (Hours) & Mem (GB) \\
\midrule
\multirow{4}{*}{\rotatebox[origin=c]{90}{\textbf{Fine-tuned}}} 
& \multirow{2}{*}{CT-ViT} & Mistral-7B & 2.00 & 61.0 & 12.00 & 76.0 \\
&  & LLaMA-2-7B & 2.00 & 62.0 & 12.00 & 73.0 \\
\cmidrule(lr){2-7}
& \multirow{2}{*}{Cosmos Tokenizer} & Mistral-7B & 1.83 & 47.5 & 11.00 & 70.0 \\
&  & LLaMA-2-7B & 1.75 & 46.5 & 11.00 & 71.6 \\

\bottomrule
\end{tabular}
}
\end{table*}

\subsection{Clinical Evaluation Metrics}
\label{appendix:clinical_eval}
\begin{figure*}[t]
    \centering
    \includegraphics[width=1.0\linewidth]{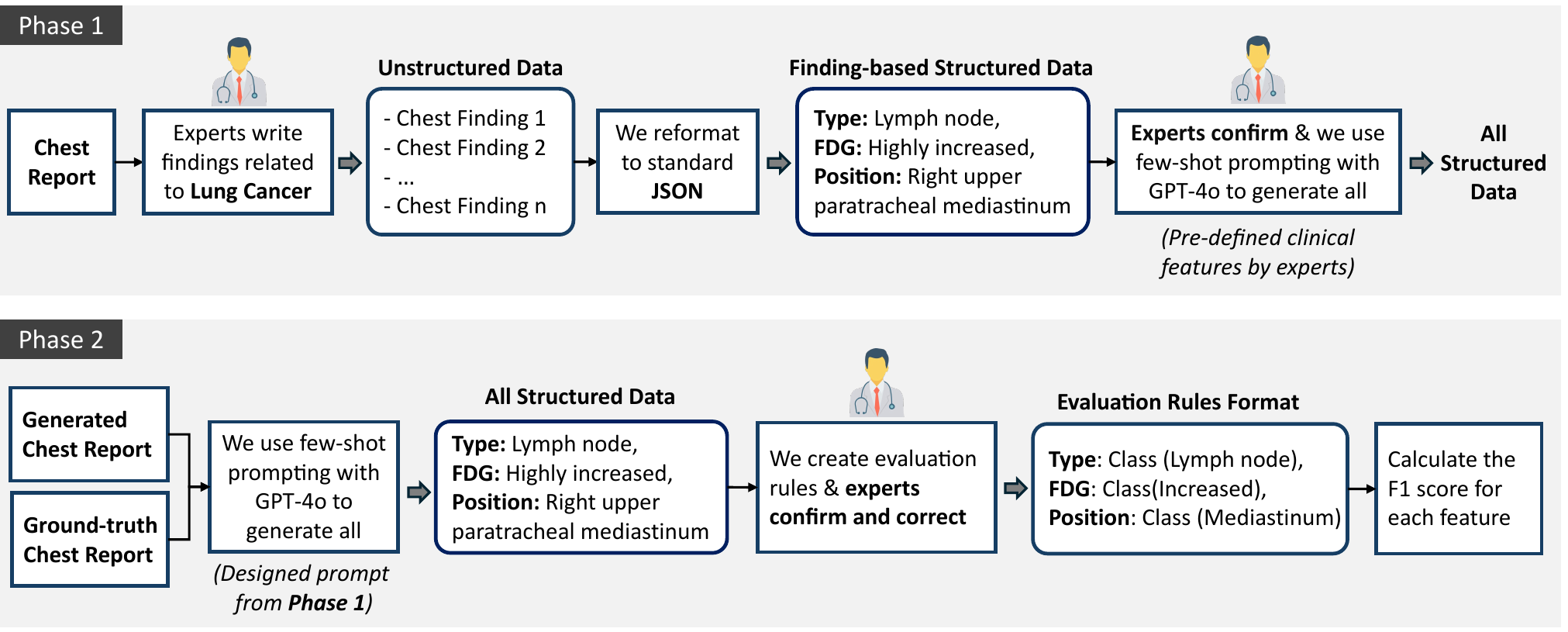}
    \caption{Clinical evaluation pipeline. {Phase 1:} Experts define structured clinical attributes from reports, which are validated and used to construct prompts for GPT-4o. {Phase 2:} GPT-4o extracts structured outputs from generated and ground-truth reports, which are mapped to clinical classes for F1-score evaluation.}
    \label{fig:B4-overall}
\end{figure*}

{
To clinically evaluate the performance of reports generated by VLMs, we propose a metric computation process developed in collaboration with medical experts. 
The overall workflow is illustrated in Figure~\ref{fig:B4-overall}, focusing on the extraction of key clinical attributes: lesion type, lesion location, and FDG uptake values.
}

\begin{figure}[th]
    \centering
    \includegraphics[width=1\linewidth]{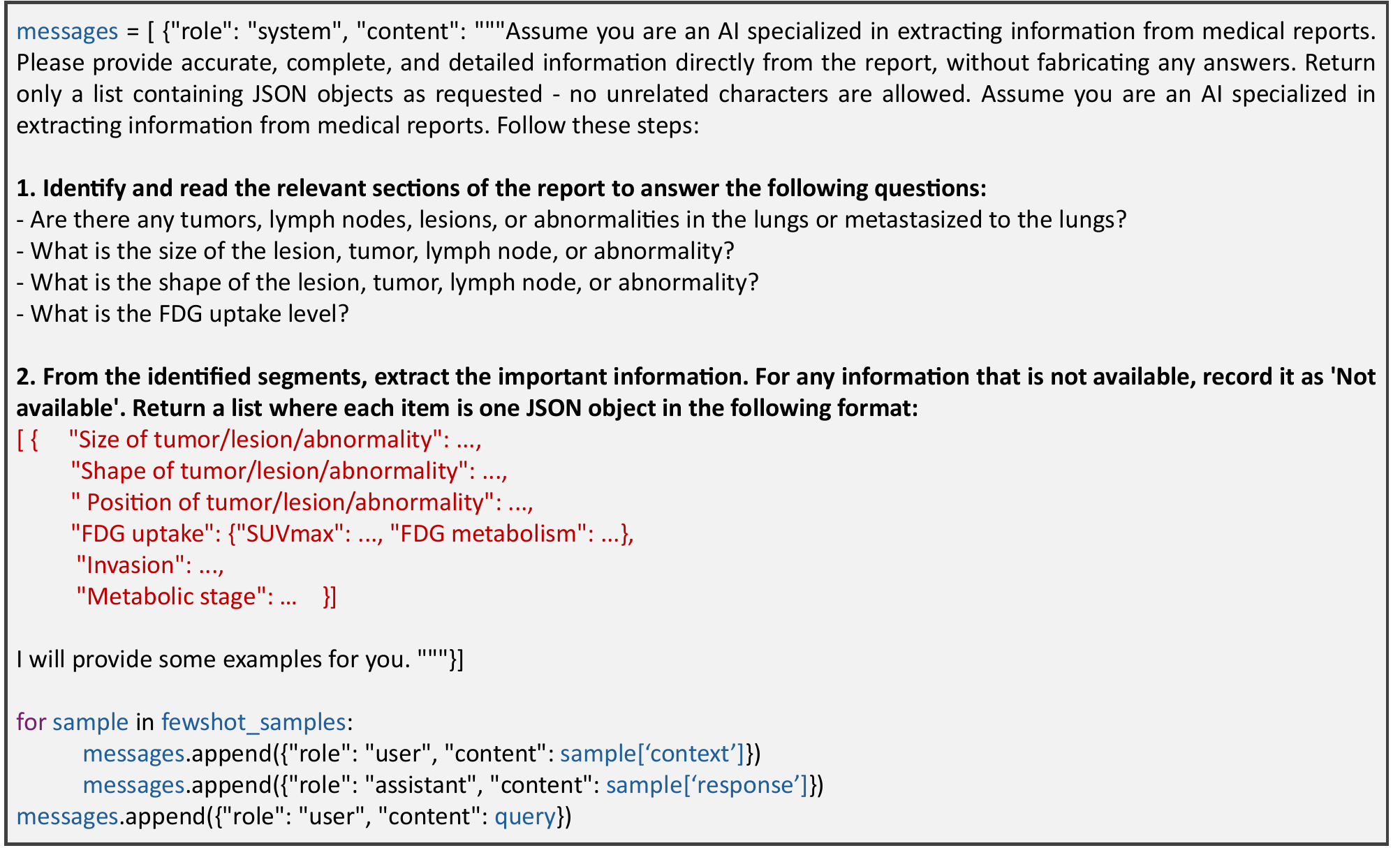}
    \caption{Message used to prompt GPT-4o for structuring VLM-generated reports into JSON format. Manually curated few-shot examples are included in the prompt, where each example consists of an input \texttt{sample[`context']} and an output \texttt{sample[`response']}. 
    See Figure~\ref{fig:fewshot_ex_extract_json} for an example.}
    \label{fig:System Message}
\end{figure}

\begin{figure*}[h]
    \centering
    \includegraphics[width=1.0\textwidth]{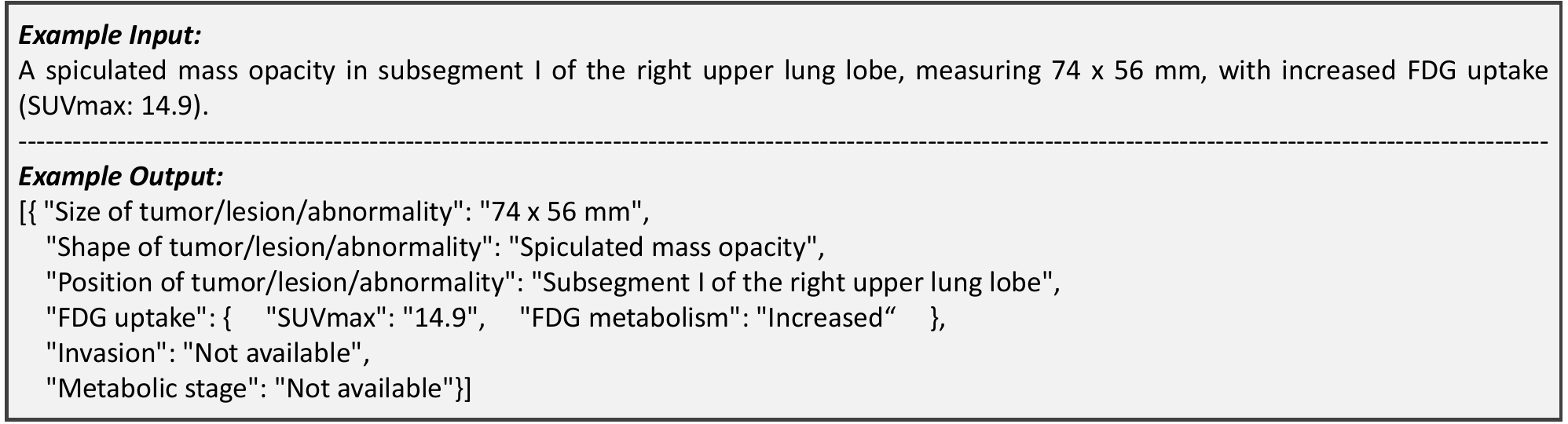}
    \caption{Example of a few-shot prompt used to guide GPT-4o in extracting structured JSON data from VLM-generated reports.}
    \label{fig:fewshot_ex_extract_json}
\end{figure*}

{
From the reports generated by the VLMs, we apply a few-shot prompting strategy with GPT-4o~\cite{hurst2024gpt} to structure the outputs, enabling systematic evaluation of each model's performance. 
The prompt used for this task is illustrated in Figures~\ref{fig:System Message} and~\ref{fig:fewshot_ex_extract_json}. 
The extracted information is subsequently mapped into categorical variables, which are validated by medical experts. 
The categories are defined as follows:
\begin{itemize}
    \item \textbf{Type}: \{lymph node, pulmonary nodule, ground-glass opacity, pulmonary mass, pleural thickening, interstitial thickening, consolidation, effusion, soft tissue nodule, wall thickening, calcified nodule, hypermetabolic lesion\}
    \item \textbf{FDG}: \{increase, not increase\}
    \item \textbf{Position}: \{mediastinum, lung, abdomen, axilla, cervical region\}
\end{itemize}
}

{
Subsequently, based on rules manually constructed in collaboration with domain experts, extracted values are grouped into semantically equivalent categories. If two values belong to the same group, they are considered equivalent for evaluation purposes. To preserve evaluation integrity, any extracted value that cannot be confidently assigned to a predefined category is labeled as {other} and excluded from positive prediction counts. 
We compute F1-scores by comparing model-generated attributes against ground truth annotations in our medical test set. The evaluation comprises four metrics: {F1-T}, which measures the F1 score based solely on lesion \textbf{T}ypes; {F1-TP}, which considers both lesion \textbf{T}ypes and \textbf{P}osition; {F1-TF}, which evaluates lesion \textbf{T}ypes together with \textbf{F}DG uptake; and {F1-TPF}, which assesses all three attributes: \textbf{T}ype, \textbf{P}osition, and \textbf{F}DG uptake.
}

\subsection{Task Evaluation by GPT-4o}

\begin{figure*}[th]
    \centering
    \includegraphics[width=\textwidth]{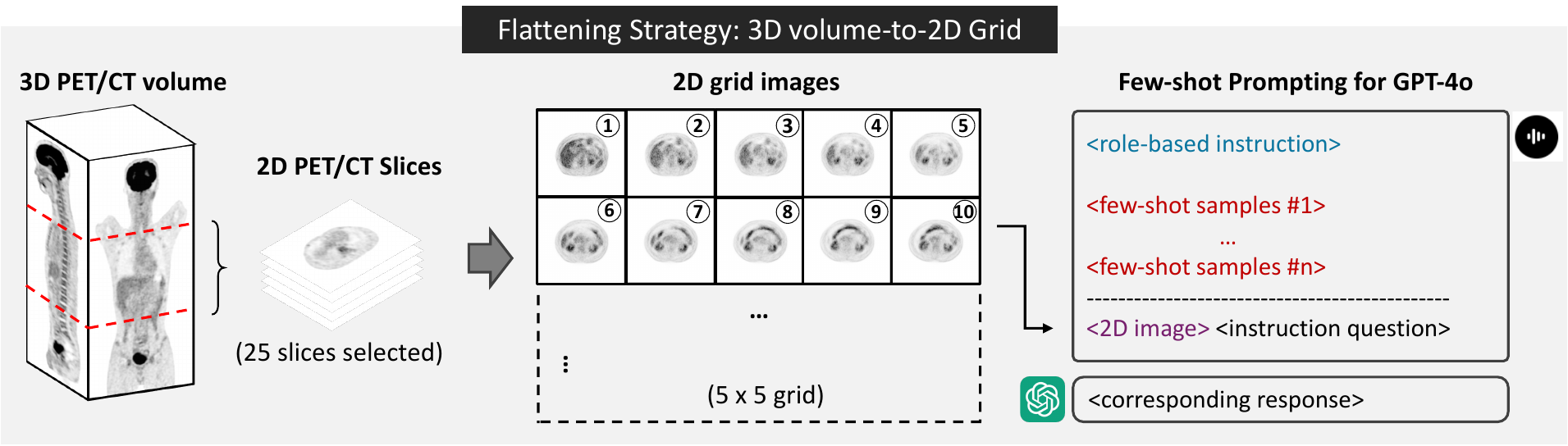}
    \caption{Visualization of the flattening strategy. Consecutive 2D slices from a 3D medical volume are arranged in numerical order (top-right corner) and concatenated into a 5×5 grid image to enable input into GPT-4o.}
    \label{fig:3d_to_2d}
\end{figure*}

\begin{figure*}[h]
    \centering
    \includegraphics[width=\textwidth]{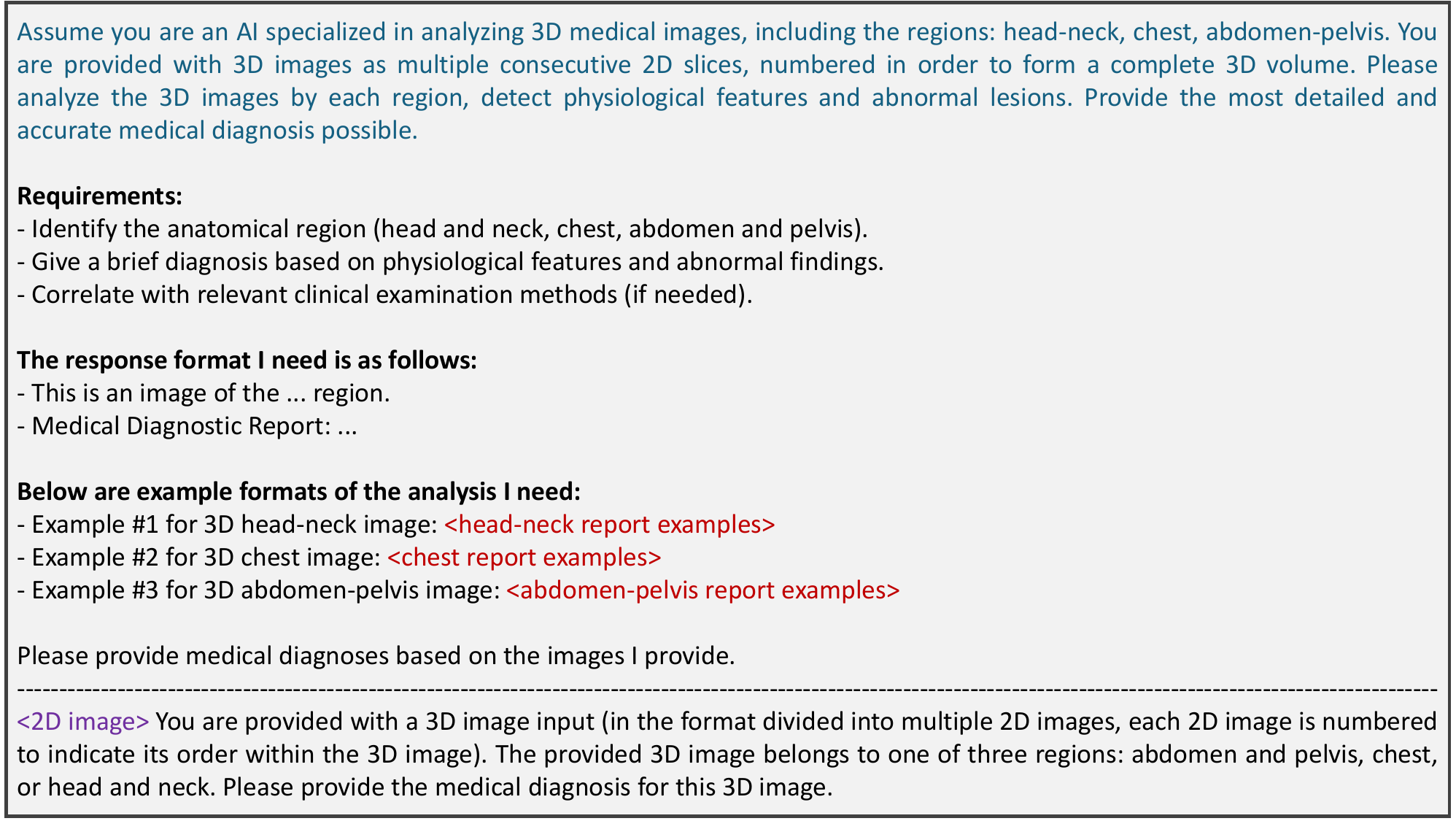}
    \caption{Prompt template used with GPT-4o to analyze concatenated 2D grid images and generate structured medical report outputs. Manually curated few-shot examples are included to guide the model.}
    \label{fig:gpt_prompt}
\end{figure*}

{
Due to the inability of GPT-4o to directly analyze 3D inputs, inspired by~\cite{wake2025open}, we adopt a flattening strategy that converts all slices of a 3D volume into multiple 2D slice images, each labeled with a numerical order in the top-right corner. 
These slices are then arranged into a 5 by 5 (25 slices) 2D grid image, as illustrated in Figure~\ref{fig:3d_to_2d}. Each 3D volume is thus represented by approximately 5 to 8 such grid images, which are subsequently input into GPT-4o with a prompt shown in Figure~\ref{fig:gpt_prompt}.
From these inputs, GPT-4o generates corresponding reports, which we then evaluate using NLP-based metrics and clinical F1 scores against the ground truth reports. 
}

\newpage
\section{Additional Results}\label{appendix:additional_res} 
\subsection{PET/CT VQA Task}
We provide additional qualitative results comparing the baseline and various fine-tuning strategies on the VQA task in Table~\ref{tab:additional_res_VQA}. 
These findings reveal several key insights that are consistent with trends observed in the report generation task.

\begin{table*}[h]
\centering
\caption{Performance on VQA task. We define training configurations as: \textbf{O}-\textbf{O}riginal dataset, \textbf{G}-Report \textbf{G}enerate dataset, \textbf{C}-Study \textbf{C}omparison dataset. R-1 and R-L denote ROUGE-1 and ROUGE-L scores. \textbf{$\uparrow$} means higher values are better. The best and second-best results are emphasized using \best{bold} and \second{underline}, respectively.
\textit{*GPT-4o is evaluated under few-shot prompting.}
} 
\label{tab:additional_res_VQA}
\resizebox{0.80\textwidth}{!}
{
\begin{tabular}{l|l|l|ccc|cccc}
\toprule
& \multicolumn{2}{c|}{\textbf{Model}} & \multicolumn{3}{c|}{\textbf{Settings}} & \multicolumn{4}{c}{\textbf{NLP Metrics $\uparrow$}} \\
\cmidrule(lr){2-3} \cmidrule(lr){4-6} \cmidrule(lr){7-10}
 & Vision & Language & O & G & C & BLEU-4 & R-1 & R-L & BERT \\
\midrule
\multirow{4}{*}{\rotatebox[origin=c]{90}{\textbf{Baseline}}} & \multicolumn{2}{c|}{LLaVA-Med~\cite{LLaVAMed2023}}  &   &  --  &    & \hspace{2pt} 3.39 & 47.83  & 33.82 & 75.86 \\
& \multicolumn{2}{c|}{M3D~\cite{M3DLaMed2022}}  &   &   -- &    & \hspace{2pt} 0.03  &  11.80 & \hspace{2pt} 9.66 &  59.87 \\
& \multicolumn{2}{c|}{RadFM~\cite{RadFM2022}}  &   &  --  &    & \hspace{2pt} 0.04 & 11.71 & 12.24 & 61.93 \\
& \multicolumn{2}{c|}{GPT-4o*~\cite{hurst2024gpt}} &   &  --  &    & \hspace{2pt} 3.01 & 49.35  & 30.09 & 71.92 \\
\midrule

\multirow{12}{*}{\rotatebox[origin=c]{90}{\textbf{Fine-tuned}}} & \multirowcell{6}[-2pt][l]{CT-ViT} & & \checkmark  & &       & 23.22 &	57.61	& 43.80 &	77.06  \\
          & & Mistral-7B & \checkmark  & \checkmark &         & \second{31.33} &	\best{65.61} &	\best{51.22} &	\best{82.05}  \\
          & & & \checkmark  & \checkmark & \checkmark    & 31.14 &	\second{65.10} &	\second{50.33} &	\second{81.80} \\
          \cmidrule(lr){3-10}
  & &    & \checkmark  & &          &  26.93	& 56.01 &	42.28 &	75.31 \\
           & & LLaMA-2-7B & \checkmark  & \checkmark &         & 26.36 &	56.48	& 42.79 &	77.73 \\
           & & & \checkmark  & \checkmark & \checkmark    & \best{31.36} &	59.14	& 48.00 &	76.72 \\
           
\cmidrule(lr){2-10}

& \multirowcell{6}[-2pt][l]{Cosmos \\ Tokenizer} &  & \checkmark  & &          & 20.01 &	58.17  &	42.54	& 76.47 \\
           & & Mistral-7B & \checkmark  & \checkmark &         & 25.71 & 61.05 &	46.59 &	78.49 \\
           & & & \checkmark  & \checkmark &  \checkmark   &   
28.09	& 62.92 &	48.37	& 79.25 \\ \cmidrule(lr){3-10}
  & &  & \checkmark  &  &          & 25.83	 & 61.80	& 46.58 &	78.87 \\
           & & LLaMA-2-7B & \checkmark  & \checkmark &         & 26.11 &	62.26 &	47.05 &	79.39 \\
           & & & \checkmark  & \checkmark & \checkmark    &  28.40 &	63.29 &	48.76	& 79.35  \\

\bottomrule
\end{tabular}
}
\end{table*}

\textbf{Comparison with existing baselines:} Fine-tuning VLMs on our proposed ViMed-PET dataset leads to significant improvements across all NLP evaluation metrics in the VQA task.

\textbf{Comparison between LLMs:} The relative performance of LLMs mirrors observations from the report generation task. When fine-tuning is limited to the original dataset (setting O), LLaMA2-7B outperforms Mistral-7B. However, with large-scale training on augmented data (settings O-G and O-G-C), Mistral-7B demonstrates superior performance, likely due to its more efficient architecture compared to LLaMA2-13B. Notably, when integrated with the Cosmos Tokenizer, LLaMA2-7B shows a modest performance advantage over Mistral-7B.

\textbf{Comparison between vision encoders:} Across all training settings, CT-ViT consistently outperforms the Cosmos Tokenizer. This indicates that CT-ViT, which is specifically designed and pretrained on 3D medical imaging data, provides greater clinical relevance and effectiveness in improving VLM performance compared to the Cosmos Tokenizer, which was pretrained on general-purpose datasets.

\subsection{Report Generation and VQA Samples}
\begin{figure*}[h]
    \centering
    \includegraphics[width=1.0\textwidth]{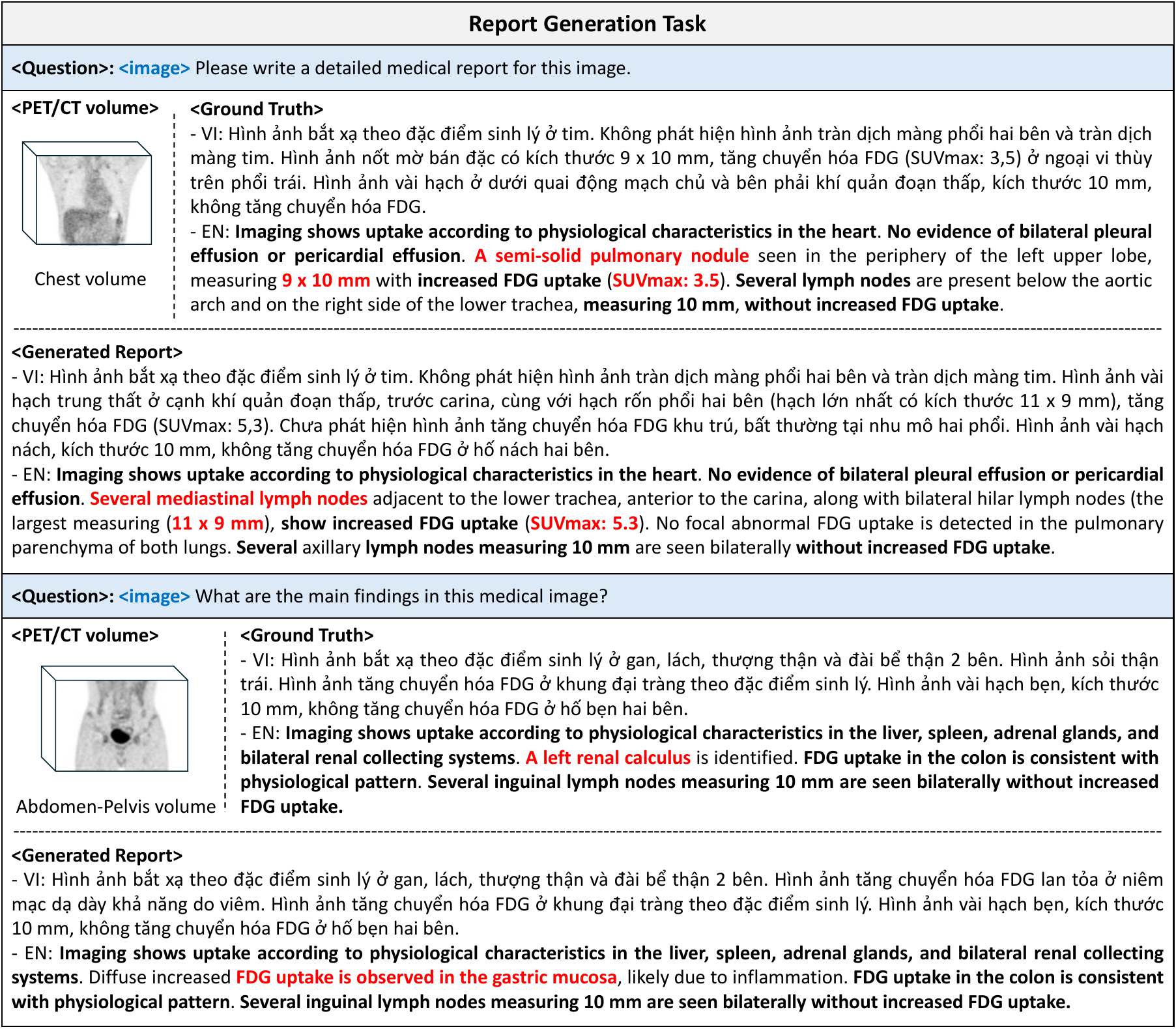}
    \caption{Ground truth and generated PET/CT reports for the chest and abdomen-pelvis regions using the CT-ViT + Mistral-7B model. Reports are originally written in Vietnamese (VI), with English (EN) translations shown for illustration. Exact matches between the generated and reference text are shown in \textbf{bold}, while key details that are missing or incorrectly added in the generated reports are marked in \textbf{\textcolor{red}{red}}.
}
    \label{fig:report_generation_samples}
\end{figure*}

\begin{figure*}[h]
    \centering
    \includegraphics[width=1\textwidth]{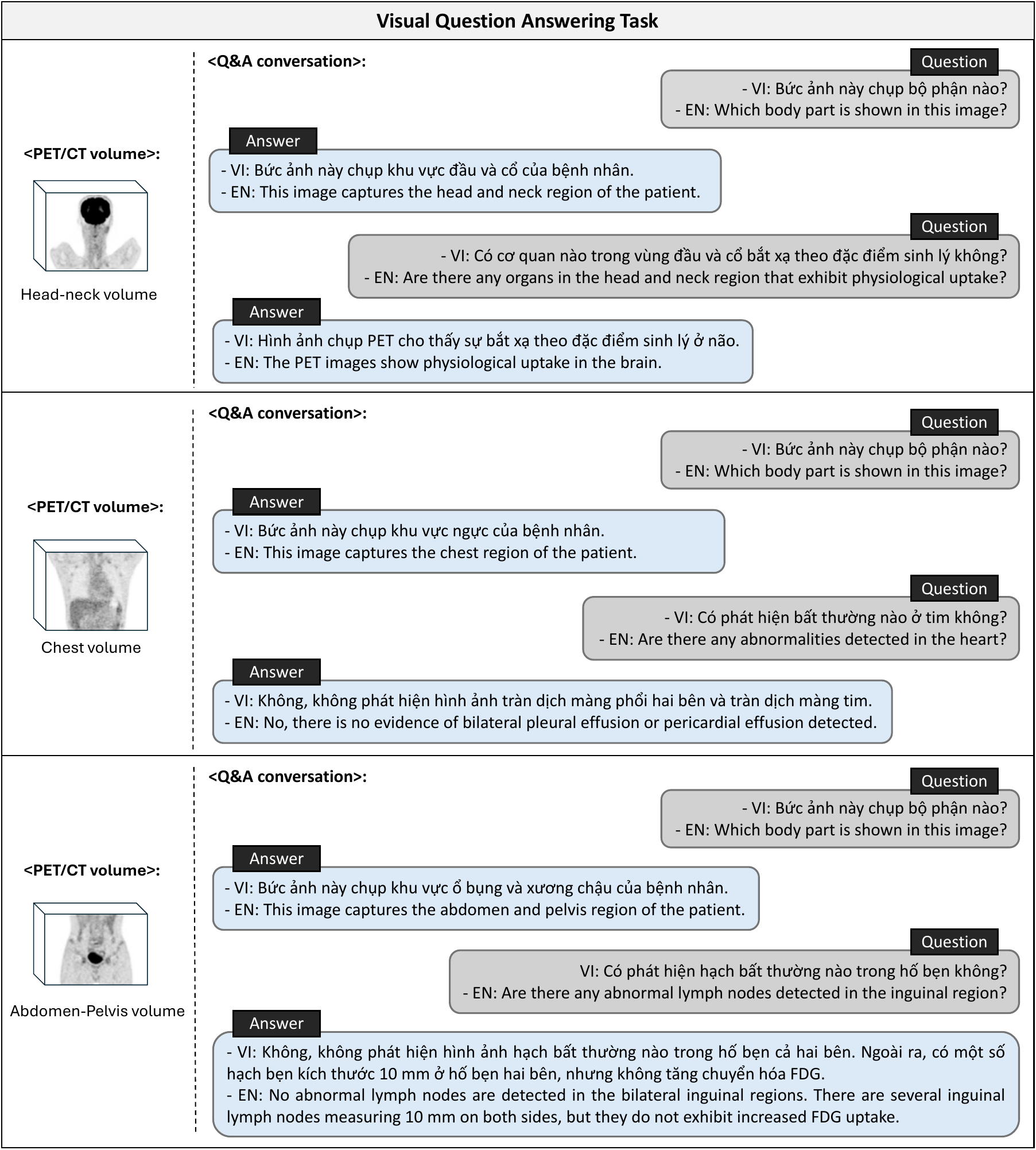}
    \caption{Short-form VQA interaction in Vietnamese (EN: translated) between a user and the CT-ViT + Mistral-7B model. The example illustrates concise factual queries and direct responses.}
    \label{fig:head_neck_vqa_samples_1}
\end{figure*}

\begin{figure*}[h]
    \centering
    \includegraphics[width=1\textwidth]{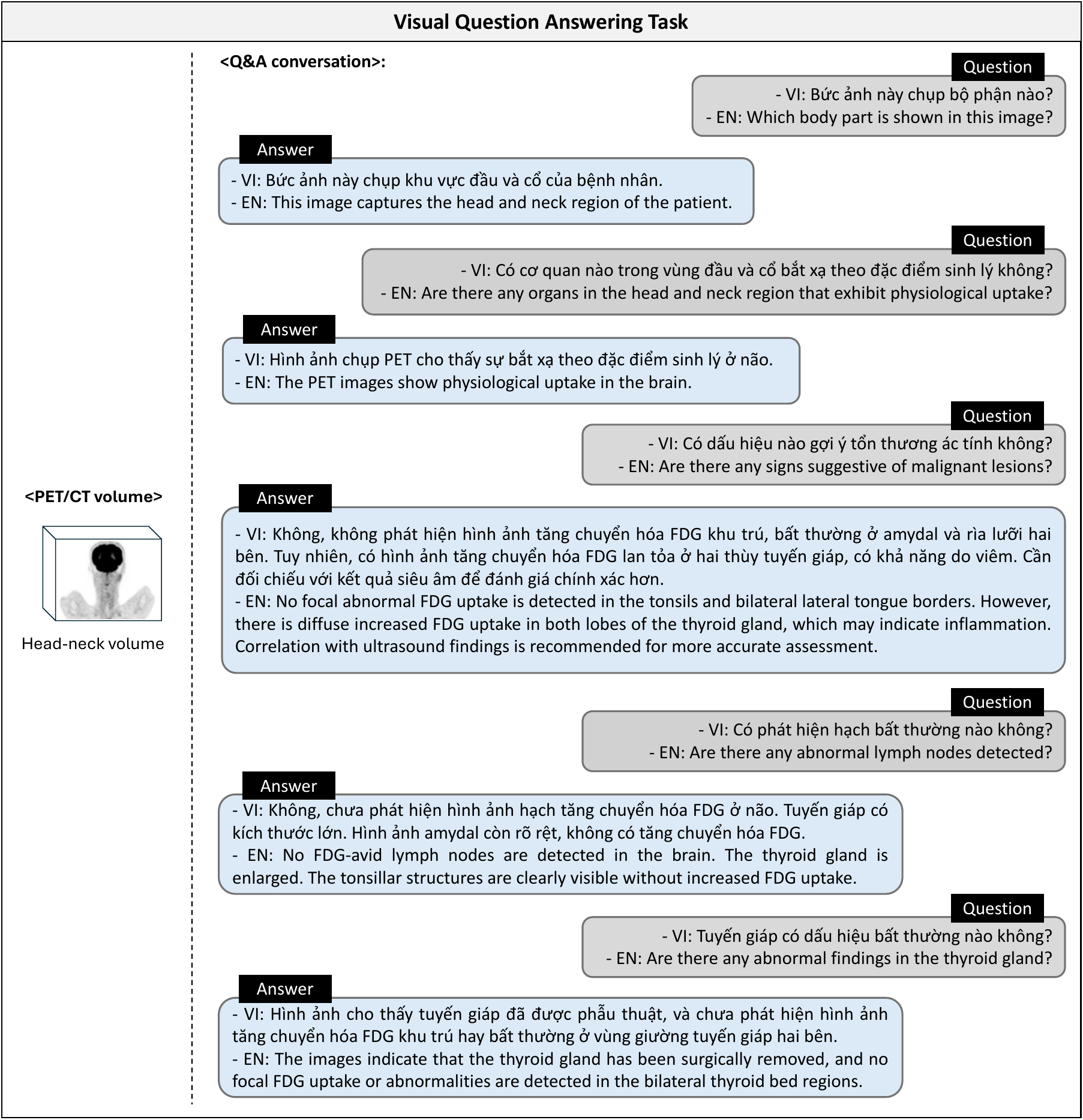}
    \caption{Long-form VQA interaction in Vietnamese (EN: translated) using the CT-ViT + Mistral-7B model. The conversation includes complex multi-sentence reasoning and detailed medical explanation.}
    \label{fig:head_neck_vqa_samples_2}
\end{figure*}

{
We present examples of generated PET/CT reports and VQA interactions using the CT-ViT + Mistral-7B combination. Figure~\ref{fig:report_generation_samples} illustrates a sample from the report generation task, highlighting both exact matches and discrepancies between predicted and ground truth reports for the chest and abdomen–pelvis regions. Additionally, Figures~\ref{fig:head_neck_vqa_samples_1} and~\ref{fig:head_neck_vqa_samples_2} show examples of VQA conversations.
}

%% file: references.bib
@article{naghavi2023cost,
  title={Cost-effectiveness of 2-[18F] FDG-PET/CT versus CE-CT for response monitoring in patients with metastatic breast cancer: a register-based comparative study},
  author={Naghavi-Behzad, Mohammad and Gerke, Oke and Kodahl, Annette Raskov and Vogsen, Marianne and Asmussen, Jon Thor and Weber, Wolfgang and Hildebrandt, Malene Grubbe and Kidholm, Kristian},
  journal={Scientific Reports},
  volume={13},
  number={1},
  pages={16315},
  year={2023},
  publisher={Nature Publishing Group UK London}
}

@article{alberts2025long,
  title={Is Long--Axial-Field-of-View PET/CT Cost-Effective? An International Health--Economic Analysis},
  author={Alberts, Ian and More, Stuart and Knapp, Karen and Mei, Riccardo and Fanti, Stefano and Mingels, Clemens and Nardo, Lorenzo and Hammond, Nii Boye and Nagaraj, Harish and Rominger, Axel and others},
  journal={Journal of Nuclear Medicine},
  year={2025},
  publisher={Society of Nuclear Medicine}
}

@article{bai2023qwen,
  title={Qwen technical report},
  author={Bai, Jinze and Bai, Shuai and Chu, Yunfei and Cui, Zeyu and Dang, Kai and Deng, Xiaodong and Fan, Yang and Ge, Wenbin and Han, Yu and Huang, Fei and others},
  journal={arXiv preprint arXiv:2309.16609},
  year={2023}
}

@article{liu2023visual,
  title={Visual instruction tuning},
  author={Liu, Haotian and Li, Chunyuan and Wu, Qingyang and Lee, Yong Jae},
  journal={Advances in Neural Information Processing Systems},
  volume={36},
  pages={34892--34916},
  year={2023}
}

@article{li2024llava,
  title={Llava-onevision: Easy visual task transfer},
  author={Li, Bo and Zhang, Yuanhan and Guo, Dong and Zhang, Renrui and Li, Feng and Zhang, Hao and Zhang, Kaichen and Zhang, Peiyuan and Li, Yanwei and Liu, Ziwei and others},
  journal={arXiv preprint arXiv:2408.03326},
  year={2024}
}

@article{bai2025qwen2,
  title={{Qwen2.5-VL} technical report},
  author={Bai, Shuai and Chen, Keqin and Liu, Xuejing and Wang, Jialin and Ge, Wenbin and Song, Sibo and Dang, Kai and Wang, Peng and Wang, Shijie and Tang, Jun and others},
  journal={arXiv preprint arXiv:2502.13923},
  year={2025}
}

@inproceedings{MedCLIP2021,
  title={Medclip: Contrastive learning from unpaired medical images and text},
  author={Wang, Zifeng and Wu, Zhenbang and Agarwal, Dinesh and Sun, Jimeng},
  booktitle={Proceedings of the 2022 Conference on Empirical Methods in Natural Language Processing.},
  volume={2022},
  pages={3876},
  year={2022}
}

@article{BiomedCLIP2021,
  title={Biomedclip: a multimodal biomedical foundation model pretrained from fifteen million scientific image-text pairs},
  author={Zhang, Sheng and Xu, Yanbo and Usuyama, Naoto and Xu, Hanwen and Bagga, Jaspreet and Tinn, Robert and Preston, Sam and Rao, Rajesh and Wei, Mu and Valluri, Naveen and others},
  journal={arXiv preprint arXiv:2303.00915},
  year={2023}
}

@article{LLaVAMed2023,
  title={Llava-med: Training a large language-and-vision assistant for biomedicine in one day},
  author={Li, Chunyuan and Wong, Cliff and Zhang, Sheng and Usuyama, Naoto and Liu, Haotian and Yang, Jianwei and Naumann, Tristan and Poon, Hoifung and Gao, Jianfeng},
  journal={Advances in Neural Information Processing Systems},
  volume={36},
  pages={28541--28564},
  year={2023}
}

@inproceedings{Med-Flamingo,
  title={Med-flamingo: a multimodal medical few-shot learner},
  author={Moor, Michael and Huang, Qian and Wu, Shirley and Yasunaga, Michihiro and Dalmia, Yash and Leskovec, Jure and Zakka, Cyril and Reis, Eduardo Pontes and Rajpurkar, Pranav},
  booktitle={Machine Learning for Health},
  pages={353--367},
  year={2023},
  organization={PMLR}
}

@article{RadFM2022,
  title={Towards generalist foundation model for radiology by leveraging web-scale 2d\&3d medical data},
  author={Wu, Chaoyi and Zhang, Xiaoman and Zhang, Ya and Hui, Hui and Wang, Yanfeng and Xie, Weidi},
  journal={Nature Communications},
  volume={16},
  number={1},
  pages={7866},
  year={2025},
  publisher={Nature Publishing Group UK London}
}

@article{M3DLaMed2022,
  title={M3d: Advancing 3d medical image analysis with multi-modal large language models},
  author={Bai, Fan and Du, Yuxin and Huang, Tiejun and Meng, Max Q-H and Zhao, Bo},
  journal={arXiv preprint arXiv:2404.00578},
  year={2024}
}

@inproceedings{Ct2rep,
  title={{CT2REP}: Automated radiology report generation for 3d medical imaging},
  author={Hamamci, Ibrahim Ethem and Er, Sezgin and Menze, Bjoern},
  booktitle={Proceedings of the 2024 International Conference on Medical Image Computing and Computer-Assisted Intervention},
  pages={476--486},
  year={2024},
  organization={Springer}
}

@article{Merlin,
  title={Merlin: A vision language foundation model for 3d computed tomography},
  author={Blankemeier, Louis and Cohen, Joseph Paul and Kumar, Ashwin and Van Veen, Dave and Gardezi, Syed Jamal Safdar and Paschali, Magdalini and Chen, Zhihong and Delbrouck, Jean-Benoit and Reis, Eduardo and Truyts, Cesar and others},
  journal={Research Square},
  pages={rs--3},
  year={2024}
}

@article{gatidis2022fdgpetctlesions,
  author       = {Gatidis, S. and Kuestner, T.},
  title        = {{A whole-body FDG-PET/CT dataset with manually annotated tumor lesions (FDG-PET-CT-Lesions)}},
  year         = {2022},
  journal  = {The Cancer Imaging Archive},
}

@article{li2020lungpetctdx,
  author       = {Li, P. and Wang, S. and Li, T. and Lu, J. and HuangFu, Y. and Wang, D.},
  title        = {A Large-Scale CT and PET/CT Dataset for Lung Cancer Diagnosis (Lung-PET-CT-Dx)},
  year         = {2020},
  journal         = {The Cancer Imaging Archive}
}

@article{vallieres2017headneckpetct,
  author       = {Martin Valli{\`e}res and Emily Kay-Rivest and L{\'e}o Jean Perrin and Xavier Liem and Christophe Furstoss and Nader Khaouam and Phuc F{\'e}lix Nguyen-Tan and Chang-Shu Wang and Khalil Sultanem},
  title        = {Data from Head-Neck-PET-CT},
  year         = {2017},
  journal         = {The Cancer Imaging Archive},
}

@article{muzi2015riderlungpetct,
  author       = {Muzi, Peter and Wanner, Matthew and Kinahan, Paul},
  title        = {RIDER Lung PET-CT: Data for Quantitative Imaging Biomarker Evaluation},
  year         = {2015},
  journal         = {The Cancer Imaging Archive}
}

@article{PLIP2021,
  title={Plip: Language-image pre-training for person representation learning},
  author={Zuo, Jialong and Hong, Jiahao and Zhang, Feng and Yu, Changqian and Zhou, Hanyu and Gao, Changxin and Sang, Nong and Wang, Jingdong},
  journal={Advances in Neural Information Processing Systems},
  volume={37},
  pages={45666--45702},
  year={2024}
}

@inproceedings{XrayGPT,
  title={XrayGPT: Chest radiographs summarization using large medical vision-language models},
  author={Thawakar, Omkar Chakradhar and Shaker, Abdelrahman M and Mullappilly, Sahal Shaji and Cholakkal, Hisham and Anwer, Rao Muhammad and Khan, Salman and Laaksonen, Jorma and Khan, Fahad},
  booktitle={Proceedings of the 23rd Workshop on Biomedical Natural Language Processing},
  pages={440--448},
  year={2024}
}

@article{ELIXR,
  title={Elixr: Towards a general purpose x-ray artificial intelligence system through alignment of large language models and radiology vision encoders},
  author={Xu, Shawn and Yang, Lin and Kelly, Christopher and Sieniek, Marcin and Kohlberger, Timo and Ma, Martin and Weng, Wei-Hung and Kiraly, Atilla and Kazemzadeh, Sahar and Melamed, Zakkai and others},
  journal={arXiv preprint arXiv:2308.01317},
  year={2023}
}

@inproceedings{lin2023pmc,
  title={Pmc-clip: Contrastive language-image pre-training using biomedical documents},
  author={Lin, Weixiong and Zhao, Ziheng and Zhang, Xiaoman and Wu, Chaoyi and Zhang, Ya and Wang, Yanfeng and Xie, Weidi},
  booktitle={International Conference on Medical Image Computing and Computer-Assisted Intervention},
  pages={525--536},
  year={2023},
  organization={Springer}
}

@article{johnson2019mimic,
  title={MIMIC-CXR, a de-identified publicly available database of chest radiographs with free-text reports},
  author={Johnson, Alistair EW and Pollard, Tom J and Berkowitz, Seth J and Greenbaum, Nathaniel R and Lungren, Matthew P and Deng, Chih-ying and Mark, Roger G and Horng, Steven},
  journal={Scientific data},
  volume={6},
  number={1},
  pages={317},
  year={2019},
  publisher={Nature Publishing Group UK London}
}

@article{ruckert2024rocov2,
  title={Rocov2: Radiology objects in context version 2, an updated multimodal image dataset},
  author={R{\"u}ckert, Johannes and Bloch, Louise and Br{\"u}ngel, Raphael and Idrissi-Yaghir, Ahmad and Sch{\"a}fer, Henning and Schmidt, Cynthia S and Koitka, Sven and Pelka, Obioma and Abacha, Asma Ben and G. Seco de Herrera, Alba and others},
  journal={Scientific Data},
  volume={11},
  number={1},
  pages={688},
  year={2024},
  publisher={Nature Publishing Group UK London}
}

@article{chen2024chexagent,
  title={Chexagent: Towards a foundation model for chest x-ray interpretation},
  author={Chen, Zhihong and Varma, Maya and Delbrouck, Jean-Benoit and Paschali, Magdalini and Blankemeier, Louis and Van Veen, Dave and Valanarasu, Jeya Maria Jose and Youssef, Alaa and Cohen, Joseph Paul and Reis, Eduardo Pontes and others},
  journal={arXiv preprint arXiv:2401.12208},
  year={2024}
}

@article{liu2023qilin,
  title={Qilin-med-vl: Towards chinese large vision-language model for general healthcare},
  author={Liu, Junling and Wang, Ziming and Ye, Qichen and Chong, Dading and Zhou, Peilin and Hua, Yining},
  journal={arXiv preprint arXiv:2310.17956},
  year={2023}
}

@article{zhang2023huatuogpt,
  title={Huatuogpt, towards taming language model to be a doctor},
  author={Zhang, Hongbo and Chen, Junying and Jiang, Feng and Yu, Fei and Chen, Zhihong and Li, Jianquan and Chen, Guiming and Wu, Xiangbo and Zhang, Zhiyi and Xiao, Qingying and others},
  journal={arXiv preprint arXiv:2305.15075},
  year={2023}
}

@inproceedings{radford2021learning,
  title={Learning transferable visual models from natural language supervision},
  author={Radford, Alec and Kim, Jong Wook and Hallacy, Chris and Ramesh, Aditya and Goh, Gabriel and Agarwal, Sandhini and Sastry, Girish and Askell, Amanda and Mishkin, Pamela and Clark, Jack and others},
  booktitle={International Conference on Machine Learning},
  pages={8748--8763},
  year={2021},
}

@article{yang2023dawn,
  title={The dawn of {LMMs}: Preliminary explorations with {GPT-4V(ision)}},
  author={Yang, Zhengyuan and Li, Linjie and Lin, Kevin and Wang, Jianfeng and Lin, Chung-Ching and Liu, Zicheng and Wang, Lijuan},
  journal={arXiv preprint arXiv:2309.17421},
  pages={1-166},
  year={2023}
}

@inproceedings{javed2024cplip,
  title={Cplip: zero-shot learning for histopathology with comprehensive vision-language alignment},
  author={Javed, Sajid and Mahmood, Arif and Ganapathi, Iyyakutti Iyappan and Dharejo, Fayaz Ali and Werghi, Naoufel and Bennamoun, Mohammed},
  booktitle={Proceedings of the IEEE/CVF 2024 Conference on Computer Vision and Pattern Recognition},
  pages={11450--11459},
  year={2024}
}

@inproceedings{yan2024ahive,
  title={AHIVE: Anatomy-aware Hierarchical Vision Encoding for Interactive Radiology Report Retrieval},
  author={Yan, Sixing and Cheung, William K and Tsang, Ivor W and Chiu, Keith and Tong, Terence M and Cheung, Ka Chun and See, Simon},
  booktitle={Proceedings of the 2024 IEEE/CVF Conference on Computer Vision and Pattern Recognition},
  pages={14324--14333},
  year={2024}
}

@inproceedings{pham2024fg,
  title={Fg-cxr: a radiologist-aligned gaze dataset for enhancing interpretability in chest x-ray report generation},
  author={Pham, Trong Thang and Ho, Ngoc-Vuong and Bui, Nhat-Tan and Phan, Thinh and Brijesh, Patel and Adjeroh, Donald and Doretto, Gianfranco and Nguyen, Anh and Wu, Carol C and Nguyen, Hien and others},
  booktitle={Proceedings of the Asian conference on computer vision},
  pages={941--958},
  year={2024}
}

@article{hurst2024gpt,
  title={Gpt-4o system card},
  author={Hurst, Aaron and Lerer, Adam and Goucher, Adam P and Perelman, Adam and Ramesh, Aditya and Clark, Aidan and Ostrow, AJ and Welihinda, Akila and Hayes, Alan and Radford, Alec and others},
  journal={arXiv preprint arXiv:2410.21276},
  year={2024}
}

@article{alayrac2022flamingo,
  title={Flamingo: a visual language model for few-shot learning},
  author={Alayrac, Jean-Baptiste and Donahue, Jeff and Luc, Pauline and Miech, Antoine and Barr, Iain and Hasson, Yana and Lenc, Karel and Mensch, Arthur and Millican, Katherine and Reynolds, Malcolm and others},
  journal={Advances in neural information processing systems},
  volume={35},
  pages={23716--23736},
  year={2022}
}

@article{moon2022multi,
  title={Multi-modal understanding and generation for medical images and text via vision-language pre-training},
  author={Moon, Jong Hak and Lee, Hyungyung and Shin, Woncheol and Kim, Young-Hak and Choi, Edward},
  journal={IEEE Journal of Biomedical and Health Informatics},
  volume={26},
  number={12},
  pages={6070--6080},
  year={2022},
  publisher={IEEE}
}

@article{nori2023can,
  title={Can generalist foundation models outcompete special-purpose tuning? case study in medicine},
  author={Nori, Harsha and Lee, Yin Tat and Zhang, Sheng and Carignan, Dean and Edgar, Richard and Fusi, Nicolo and King, Nicholas and Larson, Jonathan and Li, Yuanzhi and Liu, Weishung and others},
  journal={arXiv preprint arXiv:2311.16452},
  year={2023}
}

@article{Annotations,
  title={Towards a better understanding of annotation tools for medical imaging: a survey},
  author={Aljabri, Manar and AlAmir, Manal and AlGhamdi, Manal and Abdel-Mottaleb, Mohamed and Collado-Mesa, Fernando},
  journal={Multimedia Tools and Applications},
  volume={81},
  number={18},
  pages={25877--25911},
  year={2022},
  publisher={Springer}
}

@article{KnowledgeMatters,
  title={Knowledge matters: Chest radiology report generation with general and specific knowledge},
  author={Yang, Shuxin and Wu, Xian and Ge, Shen and Zhou, S Kevin and Xiao, Li},
  journal={Medical Image Analysis},
  volume={80},
  pages={102510},
  year={2022},
  publisher={Elsevier}
}

@article{davis2025knowledge,
  title={Knowledge-Augmented Language Models Interpreting Structured Chest X-Ray Findings},
  author={Davis, Alexander and Souza, Rafael and Lim, Jia-Hao},
  journal={arXiv preprint arXiv:2505.01711},
  year={2025}
}

@article{chambon2022roentgen,
  title={Roentgen: vision-language foundation model for chest x-ray generation},
  author={Chambon, Pierre and Bluethgen, Christian and Delbrouck, Jean-Benoit and Van der Sluijs, Rogier and Po{\l}acin, Ma{\l}gorzata and Chaves, Juan Manuel Zambrano and Abraham, Tanishq Mathew and Purohit, Shivanshu and Langlotz, Curtis P and Chaudhari, Akshay},
  journal={arXiv preprint arXiv:2211.12737},
  year={2022}
}

@article{xu2024medvilam,
  title={MedViLaM: A multimodal large language model with advanced generalizability and explainability for medical data understanding and generation},
  author={Xu, Lijian and Sun, Hao and Ni, Ziyu and Li, Hongsheng and Zhang, Shaoting},
  journal={arXiv preprint arXiv:2409.19684},
  year={2024}
}

@article{xin2025med3dvlm,
  title={Med3DVLM: An Efficient Vision-Language Model for 3D Medical Image Analysis},
  author={Xin, Yu and Ates, Gorkem Can and Gong, Kuang and Shao, Wei},
  journal={arXiv preprint arXiv:2503.20047},
  year={2025}
}

@article{nguyen2024multilingual,
  title={Multilingual diversity improves vision-language representations},
  author={Nguyen, Thao and Wallingford, Matthew and Santy, Sebastin and Ma, Wei-Chiu and Oh, Sewoong and Schmidt, Ludwig and Koh, Pang Wei W and Krishna, Ranjay},
  journal={Advances in Neural Information Processing Systems},
  volume={37},
  pages={91430--91459},
  year={2024}
}

@article{kim2016non,
  title={Non-invasive metabolic imaging of brain tumours in the era of precision medicine},
  author={Kim, Michelle M and Parolia, Abhijit and Dunphy, Mark P and Venneti, Sriram},
  journal={Nature Reviews Clinical Oncology},
  volume={13},
  number={12},
  pages={725--739},
  year={2016},
  publisher={Nature Publishing Group UK London}
}

@article{lewis2015imaging,
  title={Imaging tumor metabolism using positron emission tomography},
  author={Lewis, David Y and Soloviev, Dmitry and Brindle, Kevin M},
  journal={The Cancer Journal},
  volume={21},
  number={2},
  pages={129--136},
  year={2015},
  publisher={LWW}
}

@article{gambhir2002molecular,
  title={Molecular imaging of cancer with positron emission tomography},
  author={Gambhir, Sanjiv Sam},
  journal={Nature Reviews Cancer},
  volume={2},
  number={9},
  pages={683--693},
  year={2002},
  publisher={Nature Publishing Group UK London}
}

@article{schwenck2023advances,
  title={Advances in PET imaging of cancer},
  author={Schwenck, Johannes and Sonanini, Dominik and Cotton, Jonathan M and Rammensee, Hans-Georg and la Foug{\`e}re, Christian and Zender, Lars and Pichler, Bernd J},
  journal={Nature Reviews Cancer},
  volume={23},
  number={7},
  pages={474--490},
  year={2023},
  publisher={Nature Publishing Group UK London}
}

@inproceedings{ROUGE,
  title={Rouge: A package for automatic evaluation of summaries},
  author={Lin, Chin-Yew},
  booktitle={Text Summarization Branches Out},
  pages={74--81},
  year={2004}
}

@inproceedings{BLEU,
  title={Bleu: a method for automatic evaluation of machine translation},
  author={Papineni, Kishore and Roukos, Salim and Ward, Todd and Zhu, Wei-Jing},
  booktitle={Proceedings of the 40th Annual Meeting of the Association for Computational Linguistics},
  pages={311--318},
  year={2002}
}

@article{BertScore,
  title={Bertscore: Evaluating text generation with bert},
  author={Zhang, Tianyi and Kishore, Varsha and Wu, Felix and Weinberger, Kilian Q and Artzi, Yoav},
  journal={arXiv preprint arXiv:1904.09675},
  year={2019}
}

@article{Cosmos,
  title={Cosmos world foundation model platform for physical ai},
  author={Agarwal, Niket and Ali, Arslan and Bala, Maciej and Balaji, Yogesh and Barker, Erik and Cai, Tiffany and Chattopadhyay, Prithvijit and Chen, Yongxin and Cui, Yin and Ding, Yifan and others},
  journal={arXiv preprint arXiv:2501.03575},
  year={2025}
}

@inproceedings{CT-ViT,
  title={Generatect: Text-conditional generation of 3d chest ct volumes},
  author={Hamamci, Ibrahim Ethem and Er, Sezgin and Sekuboyina, Anjany and Simsar, Enis and Tezcan, Alperen and Simsek, Ayse Gulnihan and Esirgun, Sevval Nil and Almas, Furkan and Do{\u{g}}an, Irem and Dasdelen, Muhammed Furkan and others},
  booktitle={European Conference on Computer Vision},
  pages={126--143},
  year={2024},
  organization={Springer}
}

@article{jiang2023mistral7b,
      title={Mistral 7B}, 
      author={Albert Q. Jiang and Alexandre Sablayrolles and Arthur Mensch and Chris Bamford and Devendra Singh Chaplot and Diego de las Casas and Florian Bressand and Gianna Lengyel and Guillaume Lample and Lucile Saulnier and Lélio Renard Lavaud and Marie-Anne Lachaux and Pierre Stock and Teven Le Scao and Thibaut Lavril and Thomas Wang and Timothée Lacroix and William El Sayed},
      year={2023},
     journal={arXiv preprint arXiv:2310.06825},
}

@article{touvron2023llama,
  title={Llama 2: Open foundation and fine-tuned chat models},
  author={Touvron, Hugo and Martin, Louis and Stone, Kevin and Albert, Peter and Almahairi, Amjad and Babaei, Yasmine and Bashlykov, Nikolay and Batra, Soumya and Bhargava, Prajjwal and Bhosale, Shruti and others},
  journal={arXiv preprint arXiv:2307.09288},
  year={2023}
}

@inproceedings{nguyen2020phobert,
title     = {{PhoBERT: Pre-trained language models for Vietnamese}},
author    = {Dat Quoc Nguyen and Anh Tuan Nguyen},
booktitle = {Findings of the Association for Computational Linguistics: EMNLP 2020},
year      = {2020},
pages     = {1037--1042}
}

@article{hu2022lora,
  title={LORA: Low-rank adaptation of large language models.},
  author={Hu, Edward J and Shen, Yelong and Wallis, Phillip and Allen-Zhu, Zeyuan and Li, Yuanzhi and Wang, Shean and Wang, Lu and Chen, Weizhu and others},
  journal={International Conference on Learning Representations},
  volume={1},
  number={2},
  pages={3},
  year={2022}
}

@article{wake2025open,
  title={Open-vocabulary action localization with iterative visual prompting},
  author={Wake, Naoki and Kanehira, Atsushi and Sasabuchi, Kazuhiro and Takamatsu, Jun and Ikeuchi, Katsushi},
  journal={IEEE Access},
  year={2025},
  publisher={IEEE}
}

@article{loshchilov2017decoupled,
  title={Decoupled weight decay regularization},
  author={Loshchilov, Ilya and Hutter, Frank},
  journal={arXiv preprint arXiv:1711.05101},
  year={2017}
}

@article{loshchilov2016sgdr,
  title={Sgdr: Stochastic gradient descent with warm restarts},
  author={Loshchilov, Ilya and Hutter, Frank},
  journal={arXiv preprint arXiv:1608.03983},
  year={2016}
}

@inproceedings{chen2025q,
  title={Q-Adapter: Visual Query Adapter for Extracting Textually-related Features in Video Captioning},
  author={Chen, Junan and Nguyen, Trung Thanh and Komamizu, Takahiro and Ide, Ichiro},
  booktitle={Proceedings of the 7th ACM International Conference on Multimedia in Asia},
  pages={1--7},
  year={2025}
}
